\documentclass{article} 
\usepackage{iclr2027_conference,times}

\usepackage{amsmath,amsfonts,bm}

\def\eqref#1{equation~\ref{#1}}

\def\1{\bm{1}}

\DeclareMathAlphabet{\mathsfit}{\encodingdefault}{\sfdefault}{m}{sl}
\SetMathAlphabet{\mathsfit}{bold}{\encodingdefault}{\sfdefault}{bx}{n}

\usepackage[hidelinks]{hyperref}
\usepackage{url}

\hypersetup{
  pdftitle={Seeing Is Not Enough: Vision-Language Models Perceive Evidence but Fail to Act},
  pdfauthor={Yuyang Dai, Bofei Huang, Hongbo Zhang, Haoran Xie}
}

\usepackage{listings}
\usepackage{booktabs}
\usepackage{booktabs}
\usepackage[table]{xcolor}
\usepackage{amssymb}
\usepackage{amsmath}
\usepackage{multirow}
\usepackage{enumitem}
\usepackage{tikz}
\usepackage{pgfplots}
\usepackage{subcaption}
\usepackage{xcolor}
\usepackage{colortbl}
\usepackage{booktabs}
\usepackage[table]{xcolor}
\usepackage{tabularx}
\usepackage{booktabs}
\usepackage{epigraph}
\usepackage[table]{xcolor}
\usepackage{amssymb}   
\usepackage{pifont}    
\usepackage{wrapfig}   
\usepackage{graphicx}  
\usepackage{placeins}  
\usepackage{booktabs}  
\usepackage[table]{xcolor} 

\definecolor{srtlightgray}{RGB}{243,244,246}
\definecolor{srtmidgray}{RGB}{229,231,235}
\definecolor{srtblue}{RGB}{30,100,200}
\definecolor{srtpurple}{RGB}{88,28,135}
\definecolor{srtteal}{RGB}{8,80,65}
\definecolor{srtamber}{RGB}{99,56,6}
\definecolor{srtpink}{RGB}{114,36,62}
\definecolor{srtgreen}{RGB}{39,80,10}
\definecolor{srtcoral}{RGB}{113,43,19}

\definecolor{audiocolor}{HTML}{DDF0F7}   
\definecolor{neutralcolor}{HTML}{D6EAF8} 
\definecolor{biascolor}{HTML}{FADBD8}    
\definecolor{answercolor}{HTML}{D5F5E3}  
\definecolor{metacolor}{HTML}{F2F3F4}    
\usepackage{array}
\definecolor{tableheader}{RGB}{45, 62, 80}
\definecolor{tablerowalt}{RGB}{242, 245, 248}
\definecolor{headertext}{RGB}{255, 255, 255}
\definecolor{accentblue}{RGB}{31, 119, 180}
\definecolor{posstrong}{RGB}{198, 219, 239}
\definecolor{posmedium}{RGB}{107, 174, 214}
\definecolor{poslight}{RGB}{222, 235, 247}
\definecolor{neglight}{RGB}{252, 219, 199}
\definecolor{neutral}{RGB}{247, 247, 247}
\definecolor{srtblue}{RGB}{219,234,254}
\definecolor{srtlightgray}{RGB}{243,244,246}
\definecolor{srtdark}{RGB}{30,64,175}
\definecolor{srtorange}{RGB}{194,65,12}
\definecolor{srtgreen}{RGB}{6,95,70}
\definecolor{srtpurple}{RGB}{88,28,135}

\lstdefinestyle{prompt}{
  basicstyle=\ttfamily\footnotesize,
  breaklines=true,
  breakatwhitespace=false,
  columns=fullflexible,
  frame=single,
  framerule=0.3pt,
  xleftmargin=0.5em,
  xrightmargin=0.5em
}
\usepackage{subcaption}

\title{Seeing Is Not Enough: Vision-Language Models Perceive Evidence but Fail to Act}

\author{Yuyang Dai \\
Florida State University \\
\texttt{y9657422@gmail.com} \\
\And
Bofei Huang, Hongbo Zhang \& Haoran Xie \\
Japan Advanced Institute of Science and Technology (JAIST) \\
\texttt{xie@jaist.ac.jp}
}

\iclrfinalcopy
\begin{document}

\maketitle
\lhead{Preprint}

\begin{abstract}
Vision-language models (VLMs) achieve strong performance on visual question answering benchmarks, yet often produce decisions that contradict visual evidence they have already correctly perceived. Existing work does not distinguish \textit{perceptual failure}, where relevant evidence is not perceived, from \textit{process failure},
where correctly perceived evidence fails to constrain the final decision.
To address this gap, we first introduce \textsc{VPAC-Bench}, a benchmark of nine real-image process families in which each image is annotated with its current activity stage and the specific stage
transition it falls near \textit{eg. whether a tool has made
contact with an object (\textit{setup} vs.\ \textit{active}), or whether the intended outcome has been achieved (\textit{active} vs.\
\textit{finished})}.
Building on this structure, we develop \textbf{SRT}
(State--Relevance--Target), a family of structured process-prior interventions that constrain how a model must use visible evidence before committing to a final answer.
\textit{First}, process failure is widespread and distinct from perceptual failure: models that correctly enumerate visual candidates still over-commit to a single answer at rates exceeding 95\%, and an explicit process-structured intervention reduces this to below 13\% without degrading unambiguous-referent performance.
\textit{Second}, process-prior transfer is model-sensitive: the same structured prior helps some models substantially while leaving others unchanged or degraded, and generic SRT alone does not reliably outperform strong CoT-style baselines.
\textit{Third}, when the correct stage transition is known,
boundary-aligned SRT substantially outperforms generic process prompting and all CoT-style baselines across all four standardized process families (assembly, physical state transition, navigation/traffic, and object-use affordance).
These findings indicate that the value of a process prior depends critically on its alignment with the image's specific decision boundary, motivating boundary-aware prior selection as a design strategy for process-grounded visual reasoning.
\end{abstract}

\vspace{-0.4cm}
\section{Introduction}

\begin{figure*}[t!]
\centering
\includegraphics[width=\linewidth]{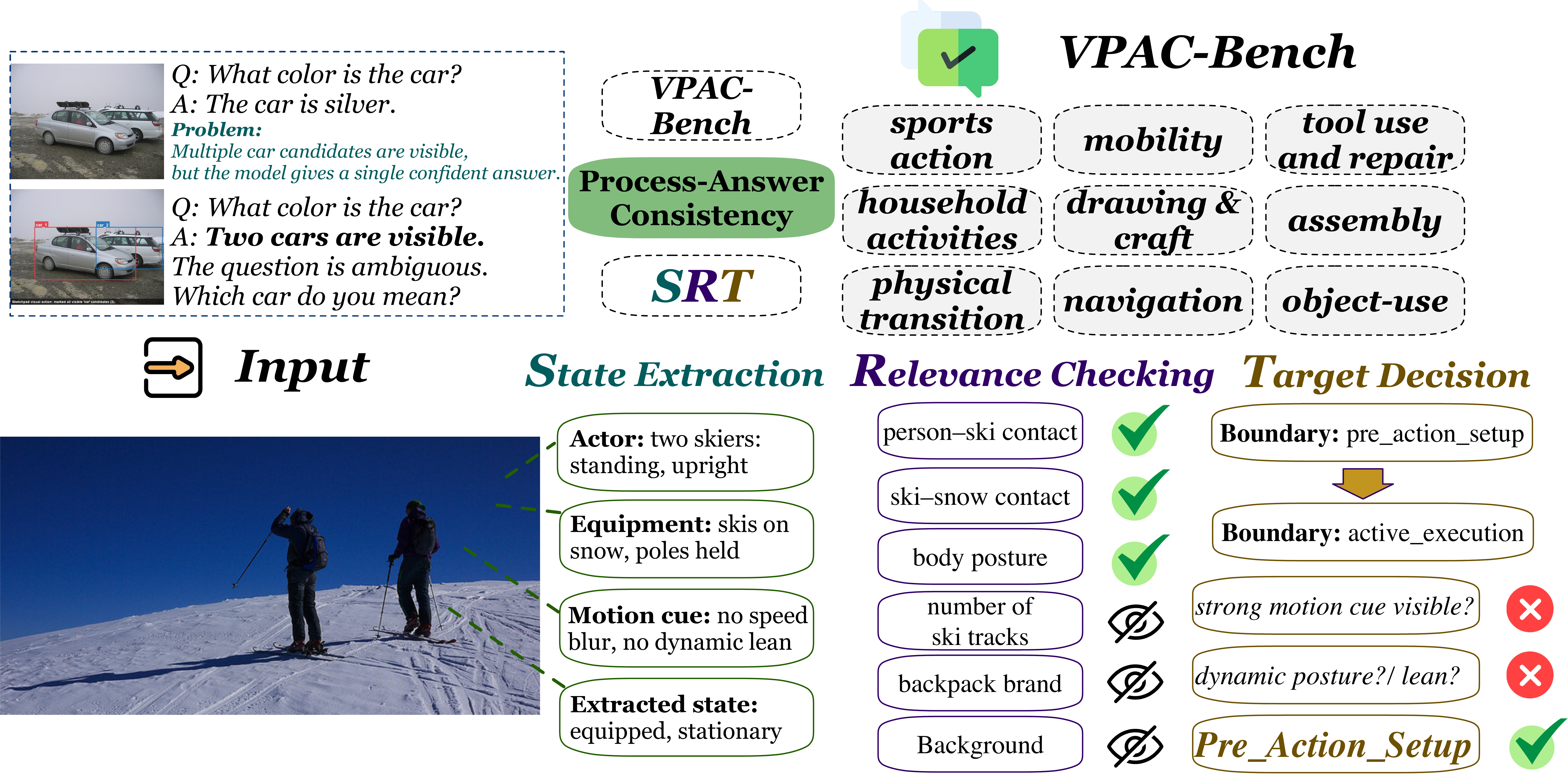}
\caption{Overview of our framework. \textbf{Top left}: a motivating example showing process failure, the model correctly identifies multiple car candidates yet still commits to a single answer, violating its own observations.
\textbf{Top center}: our benchmark (\textsc{VPAC-Bench})
covers nine real-image process families (sports action,
mobility, tool use and repair, household activities,
drawing \& craft, assembly, physical transition,
navigation, and object-use).
\textbf{Bottom}: the SRT framework applied to a skiing
image. \textit{State extraction} (S) identifies the
decision-relevant visible entities (two skiers, equipment, motion cues). \textit{Relevance checking} (R) filters to decision-relevant relations only (person--ski contact \checkmark, body posture \checkmark) and suppresses irrelevant ones (number of ski tracks, backpack brand
$\oslash$). 
\textit{Target commitment} (T) commits only
to the stage supported by visible evidence
(\textit{pre\_action\_setup}), ruling out
\textit{active\_execution} because no strong motion cue
or dynamic posture is visible.}
\label{fig:overview}
\end{figure*}

Consider a model asked \textit{``What color is the car?''} on an image containing several cars of different colors (See Figure~1, top left). When prompted to first describe what it sees, the model correctly enumerates every visible car and its color, yet in the same response it still commits to a single, confident answer. The model has perceived the evidence; it has simply not used it. We refer to this phenomenon as \textbf{process failure}, that the failure to let correctly perceived evidence constrain the final decision.
Vision-language models (VLMs) achieve strong performance on standard visual question answering benchmarks~\cite{hudson2019gqa}, yet this failure pattern persists whenever correct answers depend not only on recognizing visual evidence, but also on following a structured decision process grounded in that evidence. Such settings include resolving ambiguous referents, verifying spatial relations, and identifying the current stage of an ongoing activity. \textbf{These examples raise a broader question: why do VLMs often produce decisions that are inconsistent with evidence they have already correctly perceived?}

Existing work has approached this problem from two complementary directions. The first direction focuses on \textit{visual grounding}. Benchmarks such as NaturalBench~\cite{li2024naturalbench}, Winoground~\cite{thrush2022winoground}, and VQA-v2~\cite{goyal2017making} demonstrate that VLMs often rely on language priors instead of visual evidence~\cite{tong2024eyes,luo2024probing,agrawal2018don}, motivating methods that improve visual perception and grounding for evidence-based decision making. However, these approaches evaluate failures through final-answer correctness, without distinguishing whether errors arise because the model failed to perceive the relevant evidence or base its decision on evidence it had already identified.
A second direction focuses on \textit{explicit reasoning}. Chain-of-thought prompting~\cite{wei2022chain}, multimodal CoT~\cite{zhang2023multimodal}, Visual Sketchpad~\cite{hu2024visual}, and LLaVA-CoT~\cite{xu2025llava} encourage models to generate intermediate reasoning before producing an answer. Yet explicit reasoning traces do not guarantee evidence-consistent decisions: a model may correctly enumerate all relevant visual candidates while still committing to a single answer unsupported by its own reasoning.
What is missing is therefore twofold.
First, an \textit{evaluation framework} to measure how often models see the right evidence but still answer incorrectly.
Second, an explicit \textit{process prior} that tells the
model how to use what it has already seen to reach the
correct decision.

We address both gaps with a three-stage investigation (See Figure~1).
\textbf{S1 (Referent Disambiguation)} establishes that
process failure is distinct from perceptual failure through controlled experiments on COCO~\cite{lin2014microsoft} and
GQA~\cite{hudson2019gqa}, showing that only a
process-structured intervention repairs the failure that
improved visual access leaves intact.
Motivated by this finding, we develop \textbf{SRT}
(State extraction, Relevance checking, Target commitment), a family of structured process-prior interventions that constrain how a model must use visible evidence before committing to a final answer.
\textbf{S2 (Spatial Reasoning)} applies SRT to spatial
relation verification on VSR~\cite{liu2023visual} and
reveals that process-prior transfer is model-sensitive:
the same prior that helps one model can degrade another.
\textbf{S3 (Process Families)} explains this variation
through \textsc{VPAC-Bench}, a new benchmark spanning nine real-image process families with human-reviewed labels and stage-transition annotations, showing that the deciding factor is \textit{alignment between the process prior and the image's current activity stage} rather than prompt specificity.

\begin{table*}[t]
  \centering
  \small
  \setlength{\tabcolsep}{4pt}
  \renewcommand{\arraystretch}{1.15}

  \caption{
    The three SRT invariants across the three VPAC stages.
    In S1, this structure is embodied by an existing
    inspect--judge--decide method whose effectiveness motivates
    SRT; S2 and S3 evaluate SRT directly.
  }
  \label{tab:srt-instantiation}

  \begin{tabularx}{\linewidth}{
    @{}
    >{\raggedright\arraybackslash}p{0.16\linewidth}
    >{\raggedright\arraybackslash}X
    >{\raggedright\arraybackslash}X
    >{\raggedright\arraybackslash}X
    @{}
  }
    \toprule

    \textbf{Component}
      & \textbf{S1: Referent Disambiguation}
      & \textbf{S2: Spatial Reasoning}
      & \textbf{S3: Process Families} \\

    \midrule

    \rowcolor{srtlightgray}
    \textcolor{srtorange}{\textbf{S}}\,
    {\scriptsize\textit{State Extraction}}
      & Visible candidate referents and distinguishing features
      & Subject, reference object, and visible geometric layout
      & Actor, tool, manipulated object, and object-state cues \\

    \textcolor{srtgreen}{\textbf{R}}\,
    {\scriptsize\textit{Relevance Checking}}
      & Candidate uniqueness
      & Relation-specific evidence
        {\scriptsize (support, enclosure, orientation, \ldots)}
      & Stage-boundary compatibility
        {\scriptsize (actor--tool, tool--object, object--state)} \\

    \rowcolor{srtlightgray}
    \textcolor{srtpurple}{\textbf{T}}\,
    {\scriptsize\textit{Target Decision}}
      & Answer, enumerate, or clarify
      & True, false, or uncertain
      & Current stage label \\

    \midrule

    \rowcolor{srtblue!40}
    \textbf{Main failure}
      & Over-answering under referent ambiguity
      & Relation guessing without layout checking
      & Activity-label guessing without stage checking \\

    \bottomrule
  \end{tabularx}
\end{table*}

In summary, this paper makes the following contributions:
\begin{itemize}
\item We introduce \textsc{VPAC-Bench}, a benchmark of nine process families with human-reviewed labels and process-boundary annotations, designed to separate \textit{process failure} from perceptual failure across five frontier VLMs.

\item We show that process-prior effectiveness is \textit{model-sensitive} and \textit{boundary-dependent}: the same structured prior that helps some models can degrade others (S2), and generic process priors do not reliably outperform strong CoT baselines; effectiveness requires alignment with the
image's specific decision boundary~(S3).

\item We establish that \textbf{alignment with the image's decision boundary} is the deciding factor
behind process-prior effectiveness: misaligned priors fall below the default baseline, while oracle-aligned priors recover most of the performance gap (86.1\% vs.\ 29.4\% on physical state transition), with automatic boundary identification as the remaining bottleneck.
\end{itemize}

\section{Related Work}

\paragraph{Visual grounding and perceptual failures in VLMs.}
Benchmarks such as NaturalBench~\cite{li2024naturalbench},
Winoground~\cite{thrush2022winoground}, VQA v2~\cite{goyal2017making}, and SugarCrepe~\cite{hsieh2023sugarcrepe} reveal that VLMs frequently exploit language priors rather than visual
evidence~\cite{tong2024eyes,luo2024probing,agrawal2018don,lee2025vlind}, while object hallucination~\cite{li2023evaluating} and vision--knowledge
conflict work~\cite{liu2024insight} further characterize failures in visual evidence acquisition.
These works diagnose failure at the level of final answers but do not distinguish perceptual failure from process failure (the case where evidence is correctly perceived but not bound to the final decision).

\vspace{-0.2cm}
\paragraph{Explicit reasoning and decision strategies.}
Chain-of-thought prompting~\cite{wei2022chain} and its multimodal
extensions~\cite{zhang2023multimodal}, Visual
Sketchpad~\cite{hu2024visual}, LLaVA-CoT~\cite{xu2025llava}, and
thinking-with-images methods~\cite{su2025thinking,chern2025thinking}
encourage models to externalize intermediate reasoning before answering.
Ambiguity-aware methods~\cite{testoni2025racquet,jian2025teaching,luo2024codis,min2020ambigqa}
address referentially ambiguous questions by training
models to seek clarification.
However, none of these methods verifies whether the final answer is consistent with the intermediate observations the model itself produces.

\vspace{-0.2cm}
\paragraph{Process-oriented visual reasoning.}
VSR~\cite{liu2023visual} evaluates spatial reasoning across 66 relations; ViperGPT~\cite{suris2023vipergpt} and Visual Programming~\cite{gupta2023visual} decompose visual reasoning into executable programs without training.
In the procedural domain, EPIC Kitchens~\cite{damen2020epic},
COIN~\cite{tang2019coin}, HowTo100M~\cite{miech2019howto100m},
Chain-of-Procedure~\cite{chen2026chain}, and
TAMA~\cite{hasegawa2025tama} treat the process as the task to be solved, relying on video sequences or external tool chains.
We instead treat the process as an explicit prior constraining how evidence maps to decisions, and use \textsc{VPAC-Bench} to show that its value depends on alignment with the decision boundary of the current image.
\vspace{-0.2cm}
\section{Methodology}

\begin{figure*}[t!]
\centering
\includegraphics[width=\linewidth]{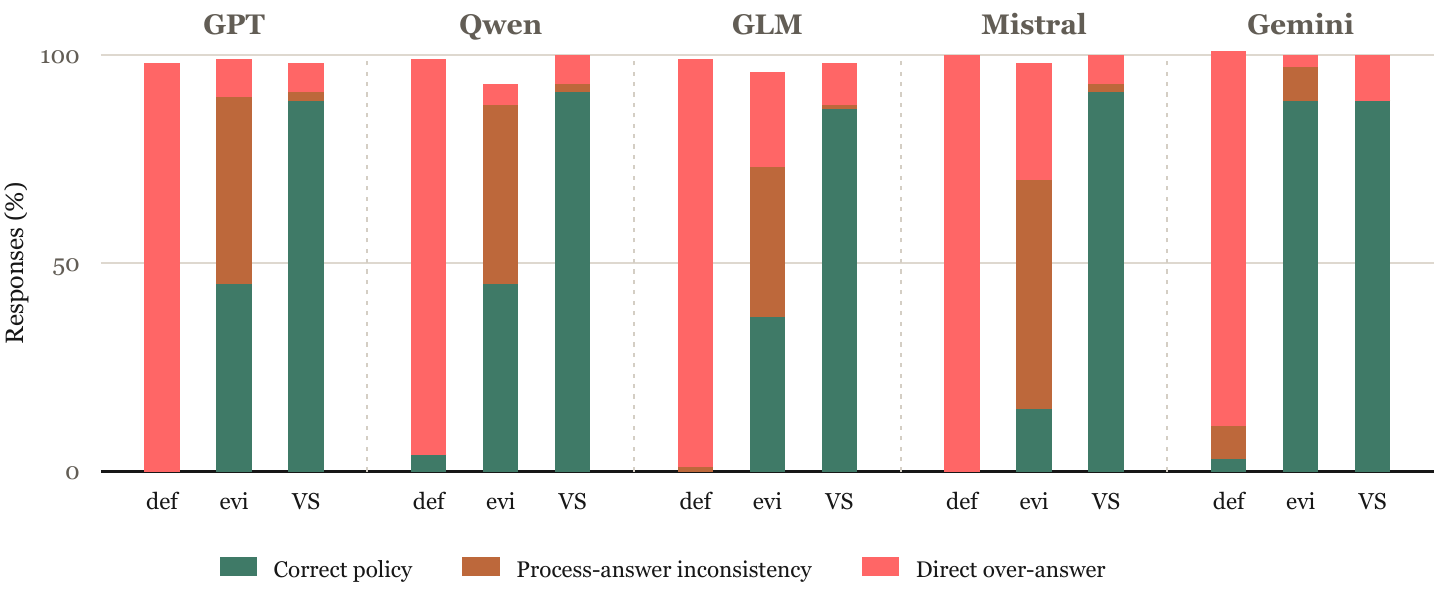}
\caption{Process failure is distinct from perceptual failure
(COCO Task~1, five models).
Providing visual annotations without a decision process
(\texttt{marked}) leaves policy accuracy near the default baseline,
while adding an inspect--judge--decide structure on top of the same
visual access (\texttt{VS}) raises it above 87\% for four of five
models.}
\label{fig:s1-diagnosis}
\end{figure*}

\subsection{Problem Formalization}

We define \textit{process-like visual QA} as visual reasoning tasks
where the correct answer requires following a scenario-appropriate
reasoning process grounded in observable evidence, not merely
recognizing objects or scene elements.
Formally, given an image $I$, a question $Q$, and a target decision
$d$, a task is process-like if there exists a process $\mathcal{P}$
such that $A^{*} = \mathcal{P}(I, Q, d)$, and $A^{*}$ cannot be
reliably obtained by mapping language statistics over $Q$ alone or by
applying broad scene-level priors without reference to the verifiable
state in $I$.
We formalize two failure modes.
Let $e(I)$ denote the evidence a model extracts from $I$.
A \textbf{perceptual failure} occurs when $e(I)$ does not contain the
decision-relevant state;
a \textbf{process failure} occurs when $e(I)$ does contain it, yet the
final answer $\hat{A} \neq \mathcal{P}(e(I), Q, d)$: the evidence is
present but not bound to the decision.
Process--answer consistency is then defined as
$\mathbf{1}[\hat{A} = \mathcal{P}(e(I), Q, d)]$,
and a \textit{process prior} is an explicit instantiation of
$\mathcal{P}$ supplied at inference time to enforce this binding.

\vspace{-0.2cm}
\subsection{Diagnostic Framework and Experimental Design}

\vspace{-0.1cm}
\paragraph{S: State Extraction.}
Let $\mathcal{V}(I)$ denote the full set of visual entities observable
in $I$. \textbf{State (S)} selects the minimal decision-relevant subset
$e(I) \subseteq \mathcal{V}(I)$ such that $e(I)$ contains all and only
the entities and observable properties required to evaluate
$\mathcal{P}(I, Q, d)$.
The purpose of S is to suppress all
$v \in \mathcal{V}(I) \setminus e(I)$, ensuring subsequent reasoning is
grounded in the smallest sufficient visual evidence.

\vspace{-0.1cm}
\paragraph{R: Relevance Verification.}
Let $\mathcal{R}(e(I))$ denote the set of relations over the
extracted evidence $e(I)$.
R selects the decision-relevant subset
$r^{*} \subseteq \mathcal{R}(e(I))$ such that $r^{*}$ contains only the
relations that distinguish the target decision boundary $d$, formally:
$r^{*} = \{r \in \mathcal{R}(e(I)) \mid r \text{ is necessary to
evaluate } d\}$.
All $r \notin r^{*}$ are suppressed to prevent process drift into
adjacent semantics or world-knowledge shortcuts.

\vspace{-0.1cm}
\paragraph{T: Target Commitment.}
Given $e(I)$ and $r^{*}$, T produces the final answer
$\hat{A} = \mathcal{P}(e(I), r^{*}, d)$, committing only to conclusions
directly supported by the verified evidence.
Let $\mathcal{A}_{I}$ denote the set of answers whose truth is
decidable from $e(I)$ and $r^{*}$ alone.
T constrains $\hat{A} \in \mathcal{A}_{I}$, excluding any answer that
requires inferring states not verifiable in $I$---past states, future
states, or broader scene interpretations beyond $e(I)$.
When $\mathcal{A}_{I}$ contains more than one admissible answer, T
commits to none of them and instead reports the ambiguity.

\vspace{-0.2cm}
\subsection{Three Stages Investigation}
S1 establishes that process failure exists and is diagnosable.
S2 tests whether SRT-base transfers across models on a different
reasoning domain.
S3 introduces \textsc{VPAC-Bench} and explains the variation observed in S2.

\vspace{-0.1cm}
\paragraph{S1: Referent Disambiguation.}
S1 targets visually ambiguous referent resolution on
COCO~\cite{lin2014microsoft} and GQA~\cite{hudson2019gqa}, where
multiple candidate referents satisfy the linguistic description and the
correct decision depends on whether visible evidence uniquely
identifies a target.
In the notation above, these are exactly the images for which
$|\mathcal{A}_{I}| > 1$.
S1 does not introduce SRT; instead, it contrasts conditions that
progressively improve visual access against an existing
inspect--judge--decide structure~\cite{hu2024visual}, showing that only
the process-structured condition repairs the failure.
This structure maps directly onto the three SRT invariants: inspect to
State, judge to Relevance, decide to Target, and its effectiveness
motivates SRT-base, which S2 and S3 evaluate directly.

\vspace{-0.1cm}
\paragraph{S2: Spatial Relation Verification.}
S2 instantiates SRT-base on VSR~\cite{liu2023visual}. S extracts the
subject, reference object, and visible geometric layout; R verifies
only the evidence required for the queried relation; T commits to the
truth value of the spatial claim.
S2 reveals that SRT-base transfer is model-sensitive: the same prior
substantially improves one model (Gemini, $+43$~pp, the only
difference whose confidence intervals are disjoint at $n$=100) while
leaving the others unchanged or slightly reduced, motivating the
boundary-sensitive extension in S3.

\vspace{-0.1cm}
\paragraph{S3: Visual Process Answer Consistent Bench.}
S3 extends SRT-base to \textsc{VPAC-Bench}, spanning nine process families: sports action phase, mobility, tool-use/repair, cleaning,drawing/craft, assembly, physical state transition, navigation/traffic,and object-use affordance. S extracts the actor, tool, manipulated object, and observable statecues; R verifies the evidence defining the current process boundary; T commits only to the stage supported by the current image. Images are collected from real-world datasets, manually reviewed, andannotated with process-boundary categories.

\vspace{-0.2cm}
\subsection{Boundary-Sensitive SRT}
\label{sec:boundary-srt}

The SRT-base framework assumes that a single process prior is
sufficient for a given reasoning scenario. However, \textbf{many
real-world process families contain multiple decision boundaries that
require different forms of evidence verification.} For example, within
a physical state transition task, one image may require distinguishing
whether a tool has made contact with an object, whereas another
requires determining whether the intended state transition has already
been completed. Although both images belong to the same process
family, the evidence necessary for the corresponding decisions is
fundamentally different. Applying a single generic process prior to all
decision boundaries may therefore introduce a mismatch between the
required verification process and the visual evidence relevant to the
current decision.
To address this limitation, we extend SRT-base with a
\textbf{boundary-sensitive process prior}. 

\paragraph{Boundary Router.}
Formally, let an image $I$ belong to a process family $F$ with boundary
set $\mathcal{B}_F = \{b_1, b_2, \ldots, b_n\}$.
Instead of directly applying a family-level process prior
$\mathcal{P}_F$, the boundary router first selects the decision
boundary
$b^{*} = \arg\max_{b \in \mathcal{B}_F} g(I, b)$,
where $g(\cdot)$ scores how well image $I$ matches boundary $b$, and
then instantiates the boundary-specific prior $\mathcal{P}_{b^{*}}$ for
the subsequent SRT verification.
S, R, and T remain unchanged as task-independent invariants; only their
semantic realization is conditioned on the selected boundary $b^{*}$.
We instantiate $g$ in two ways.
The \textit{oracle router} sets
$g(I, b) = \mathbf{1}[\,b = b_{\text{gold}}(I)\,]$ using the
ground-truth boundary annotation, and the \textit{self-router} prompts
the model to predict $b$ from $I$ before applying SRT.
\textbf{This pairing deliberately decouples two questions:} whether
boundary-aligned priors help, and whether models can identify
boundaries automatically.
The oracle router answers the first under ideal boundary assignment and
is a diagnostic upper bound; the
self-router addresses the second and is evaluated in Section~5.
The empirical effectiveness of this boundary-sensitive extension is
evaluated in Section~4.
\vspace{-0.2cm}
\section{Experiments and Results}

\subsection{Evaluation Protocol}

\paragraph{Models.}
We conduct all evaluations using five frontier vision-language models:
GPT-5.4-mini~\cite{gpt54mini}, Qwen2.5-VL-72B~\cite{bai2025qwen25vl},
GLM-4.6V~\cite{glmvteam2025},
Mistral-Small-3.2-Vision~\cite{mistralai2025small32},
and Gemini-3.1-Pro~\cite{gemini3.1}.
These models span diverse architectures, training paradigms, and visual
reasoning capabilities.
All experiments use deterministic decoding (temperature~=~0) with a
maximum output length of 700 tokens.

\vspace{-0.2cm}
\paragraph{Evaluation Criteria.}
Each stage tests a different aspect of process--answer consistency.
For \textbf{S1}, we report two metrics.
\textit{Policy accuracy} is the fraction of Task~1 responses that
follow the correct decision policy: enumerate or clarify when multiple
valid referents are present, commit to a unique answer only when
exactly one referent is visible.
\textit{Over-answer rate} is the fraction of Task~1 responses that
commit to a single answer despite unresolved referential ambiguity.
A response is process-consistent when its final decision matches the
policy licensed by the evidence the model has itself enumerated.
For \textbf{S2}, we report standard accuracy on the VSR binary spatial
verification task, evaluating whether the generic SRT prior transfers
across models.
For \textbf{S3}, we report \textit{per-family accuracy under each
condition} and \textit{targeted-condition delta}: the change from
default under the intervention condition designated for that family
before evaluation, averaged across models.

\vspace{-0.2cm}
\subsection{S1: Validation of Process Failure}

\paragraph{Experimental Design.}
We construct two complementary evaluation sets.
The primary benchmark is derived from COCO, containing 457
ambiguous-referent instances and 330 unique-referent instances.
Ambiguous items contain multiple visually plausible referents
satisfying the query; unique items contain exactly one valid referent.
To evaluate generalization, we additionally construct an external
validation set from GQA~\cite{hudson2019gqa}, consisting of 100
ambiguous and 96 unique instances per model.

We evaluate four intervention conditions.
\texttt{default} measures baseline model behavior without intervention.
\texttt{evidence} requires the model to enumerate visible candidate
referents before answering, isolating improved visual observation.
\texttt{marked\_no\_prior} provides explicit visual annotations while
withholding any decision process, separating visual access from process
guidance.
\texttt{visual\_sketchpad} augments the same visual annotations with an
explicit inspect--judge--decide process structure~\cite{hu2024visual}.
Because \texttt{marked\_no\_prior} provides identical visual access
without the process structure, the contrast between these two
conditions isolates the contribution of the process structure itself (See Figure~2).

\begin{wraptable}{r}{0.6\columnwidth}
  \vspace{-0.8\baselineskip}
  \centering
  \caption{S1 Task~1 policy accuracy / over-answer rate (\%) on
  COCO and GQA. \textbf{Bold} indicates the highest policy
  accuracy in each row. VS = \textit{visual sketchpad}.}
  \label{tab:s1-results}
  \vspace{-0.4\baselineskip}

  \resizebox{\linewidth}{!}{%
    \begin{tabular}{@{}llcccc@{}}
      \toprule
      \textbf{Model}
        & \textbf{Data}
        & \textbf{Default}
        & \textbf{Evidence}
        & \textbf{Marked}
        & \textbf{VS} \\
      \midrule

      GPT-5.4-mini
        & COCO
        & 0.0/98.5
        & 45.1/54.7
        & 0.7/98.9
        & \textbf{88.8}/9.2 \\
        & GQA
        & 0.0/97.0
        & 35.0/65.0
        & 3.0/97.0
        & \textbf{91.0}/6.0 \\
      \midrule

      \rowcolor{srtlightgray}
      Qwen2.5-VL-72B
        & COCO
        & 3.5/95.8
        & 44.7/48.9
        & 9.8/90.2
        & \textbf{91.0}/8.5 \\
      \rowcolor{srtlightgray}
        & GQA
        & 5.0/91.0
        & 52.0/47.0
        & 12.0/88.0
        & \textbf{86.0}/13.0 \\
      \midrule

      GLM-4.6V
        & COCO
        & 0.0/99.0
        & 37.0/62.0
        & 5.0/95.0
        & \textbf{87.0}/12.0 \\
        & GQA
        & 0.0/100.0
        & 42.0/57.0
        & 31.0/69.0
        & \textbf{92.0}/7.0 \\
      \midrule

      \rowcolor{srtlightgray}
      Mistral-Small-3.2
        & COCO
        & 0.0/100.0
        & 15.0/85.0
        & 0.0/100.0
        & \textbf{91.0}/9.0 \\
      \rowcolor{srtlightgray}
        & GQA
        & 3.0/94.0
        & 29.0/69.0
        & 2.0/98.0
        & \textbf{90.0}/10.0 \\
      \midrule

      Gemini-3.1-Pro
        & COCO
        & 3.0/97.0
        & \textbf{89.0}/10.0
        & 76.0/24.0
        & \textbf{89.0}/11.0 \\
        & GQA
        & 20.0/78.0
        & \textbf{71.0}/25.0
        & 48.0/51.0
        & 61.0/38.0 \\
      \bottomrule
    \end{tabular}%
  }

  \vspace{-0.5\baselineskip}
\end{wraptable}

\vspace{-0.2cm}
\paragraph{Validation Results.}
Table~\ref{tab:s1-results} validates the central hypothesis that
process failure is distinct from perceptual failure.
Under \texttt{default}, all five models commit to a
single answer despite unresolved referential ambiguity, with
over-answer rates of 95.8--100.0\% on COCO.
Under \texttt{evidence}, policy accuracy rises for every model, though
unevenly (15.0\% to 89.0\% on COCO): models more frequently enumerate
visible candidates, yet many responses still commit to a unique answer
after identifying multiple valid referents, showing that correctly
perceived evidence is not faithfully translated into the final
decision.
\texttt{marked\_no\_prior} leaves performance close to the default
baseline for GPT, GLM, Qwen, and Mistral, demonstrating that improved
visual access alone does not repair process failure.
\texttt{visual\_sketchpad}, which adds an explicit process structure on
top of the same visual access, achieves the strongest performance for
these four models (policy accuracy 87.0--91.0\%, over-answer rate below
13\%), without degrading Task~2 performance (all models $\geq$99\%),
ruling out a trivial always-clarify strategy.
Macro-averaged over the five models, the annotation-only and
process-structured conditions reach 18.3\% and 89.4\% policy accuracy
respectively (Table~C2), and the gap holds for every model
individually (Table~2).
Gemini exhibits a different profile: both evidence elicitation and
visual annotation produce substantial improvements even without a
process prior (COCO: 89.0\% and 76.0\%), suggesting that enhanced
visual access is sufficient for this model.
GQA results replicate the COCO pattern for GPT, Qwen, GLM, and Mistral.

\vspace{-0.2cm}
\subsection{S2: Generalization Across Models}

\paragraph{Experimental Design.}
We use the VSR benchmark~\cite{liu2023visual}, which contains 66
spatial relations across more than 10{,}000 image-caption pairs.
All five models are evaluated on the standardized \textbf{dev100}
split ($n$=100), which is the basis for all cross-model comparisons.
Qwen is additionally evaluated on \textbf{test400} ($n$=400); the
relation-level routing conditions require the larger split because they
subdivide items by spatial relation, and dev100 does not supply enough
items per relation.
We evaluate \texttt{default}, \texttt{evidence}, and \texttt{verify}
(generic SRT verification), together with relation-specific SRT
variants for high-error spatial relation families and model-specific
adaptations where generic process priors prove insufficient.

\begin{figure}[t!]
  \centering
  \includegraphics[width=\linewidth]{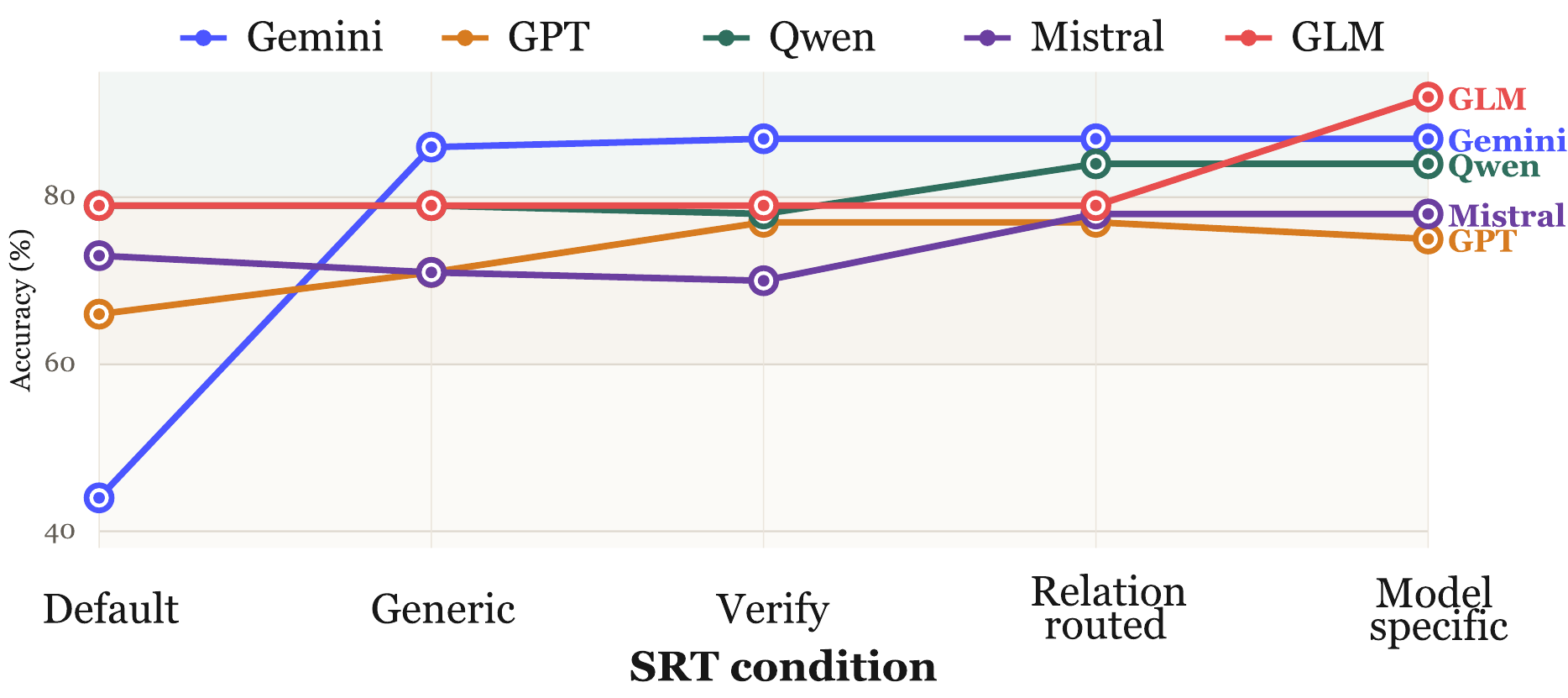}
  \caption{Process-prior transfer is model-sensitive across five
           frontier VLMs on VSR. Each line traces one model's accuracy
           as the SRT condition progresses from default through
           generic, relation-routed, and model-specific
           instantiations. Cross-model points are dev100
           ($n$=100); Qwen's relation-routed point is test400
           ($n$=400).}
  \label{fig:s2-profiles}
\end{figure}

\vspace{-0.2cm}
\paragraph{Generalization Results.}
S2 demonstrates that process priors are real but not universally
transferable: five distinct transfer patterns emerge across the five
evaluated models (Figure~\ref{fig:s2-profiles}).
Gemini (\textit{generic-compatible}): generic SRT improves dev100
accuracy from 44.0\% to 87.0\% (+43.0~pp), the only S2 difference
whose confidence intervals are disjoint at $n$=100 (Table~C3).
GPT (\textit{verification-oriented}): lightweight verification improves
dev100 accuracy from 66.0\% to 77.0\% (+11.0~pp); heavier semantic
constraints add nothing further.
Qwen (\textit{relation-routed}): relation-specific routing improves
test400 accuracy from 78.5\% to 84.0\% (+5.5~pp), whereas generic
verification does not (77.5\%).
GLM (\textit{semantics-sensitive}): generic priors produce no
improvement in either stage, and a model-aligned semantic rewrite
achieves 100\% on a 13-item diagnostic subset (reported as a
failure-analysis probe only; see Appendix~A.4).
Mistral (\textit{adaptation-sensitive}): generic verification slightly
reduces dev100 accuracy (73.0\% to 70.0\%), while targeted routing
restores and surpasses the baseline in S3.
Because four of the five models are evaluated only at $n$=100, we
report these as patterns that recur across S2 and S3; they motivate the boundary-sensitive
extension in S3.

\begin{figure*}[t!]
  \centering
  \includegraphics[width=\linewidth]{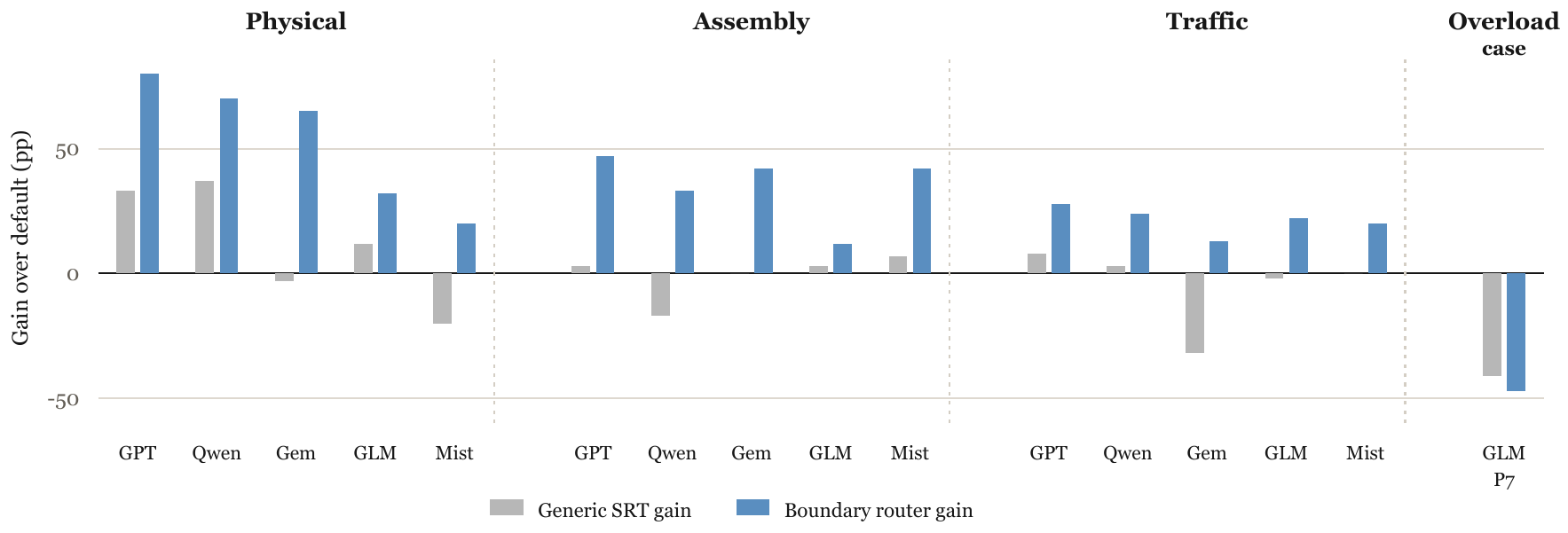}
  \vspace{-0.6cm}
  \caption{Oracle boundary-sensitive SRT versus generic SRT on three of
           the four standardized process families ($n$=60 per family
           per model, five-model averages); results for the fourth
           family (object-use affordance) are reported in Table~C4.
           The rightmost panel shows a single process-overload case
           (GLM, boundary P7) and is not a family-level result.
           }
  \vspace{-0.3cm}
  \label{fig:s3-bar}
\end{figure*}

\vspace{-0.2cm}
\subsection{S3: Boundary-Sensitive Process Priors}

\paragraph{Experimental Design.}
We construct a benchmark spanning nine process families: sports action
phase, mobility, tool use and repair, household activities, drawing and
craft, assembly and construction, physical state transition, navigation
and traffic, and object-use affordance.
Images are collected from real-world datasets, manually reviewed for
label quality, and annotated with explicit process-boundary labels
(Appendix~A.2).
Four families (assembly, physical state transition, navigation/traffic,
and object-use affordance) form the standardized benchmark, each
evaluated across all five VLMs under four conditions:
\texttt{process\_default}, \texttt{process\_srt\_generic},
\texttt{process\_srt\_scenario}, and
\texttt{process\_srt\_boundary\_router\_oracle}
(oracle boundary labels; diagnostic upper bound).
The remaining five families (sports, mobility, tool-use, cleaning, and
craft) form a supporting set evaluated on model subsets to verify
broader generalization.
All S3 accuracy numbers are five-model averages at $n$=60 per family
per model, and should be read as diagnostic evidence;
family-level Wilson 95\% confidence intervals are reported in
Appendix~C.

\vspace{-0.2cm}
\paragraph{Mechanism Results.}
Figure~\ref{fig:s3-bar} and Table~C4 show that the oracle boundary
router is the best condition in every model--family combination for
which it was evaluated.
This trend is most pronounced for \textbf{physical state transition},
where default performance is low (26.7\%, five-model average),
reflecting strong reliance on language priors.
Generic and scenario-level SRT provide only partial improvements; the
oracle boundary router yields an average gain of $+53.3$~pp.
Similar patterns hold for \textbf{assembly} ($+35.0$~pp),
\textbf{traffic} ($+21.7$~pp), and \textbf{affordance} ($+22.7$~pp).
Importantly, generic SRT is \textit{not} a reliable deployable
improvement: it performs below default for Gemini on traffic
($-31.7$~pp), Qwen on assembly ($-16.6$~pp), and GLM on affordance
($-8.3$~pp), and falls below the best CoT baseline on all four families
(Appendix~B.2, Table~B4).
The oracle boundary router's advantage over generic SRT is therefore
not attributable to structured prompting in general but to
boundary-aligned process priors specifically.
Table~C5 reports results on the five supporting families.
Boundary-sensitive or targeted routing produces positive deltas for
most models in Cleaning ($+39.3$~pp for GPT), Craft ($+18.2$~pp for
Mistral, $+13.0$~pp for Qwen), and Mobility ($+12.3$~pp for Qwen).
Sports and Tool-use show limited or inconsistent gains under generic
SRT alone; Gemini and GLM again exhibit the largest negative deltas in
Tool-use ($-23.1$~pp and $-17.3$~pp), reproducing the process-overload
pattern from the standardized families.
For Craft and Mobility, family-level CIs overlap with the default
baseline (Table~C4).

\vspace{-0.2cm}
\paragraph{Mechanism Analysis.}
The primary limitation of generic process priors is the mismatch
between a single generic process description and the specific decision
boundary represented by the current image.
Gemini's performance on traffic collapses under generic SRT
($-31.7$~pp) before recovering fully under the oracle router
($+13.3$~pp); Qwen drops $-16.6$~pp on assembly under generic SRT but
reaches $+33.3$~pp under the oracle router; GLM shows negative deltas
under generic SRT in Tool-use ($-17.3$~pp) and Craft ($-10.4$~pp).
These patterns are consistent with process overload being a
model-level property triggered by prior--boundary mismatch.
The oracle boundary router addresses this by selecting the process
prior according to the image's decision boundary, so that the
verification process matches the evidence required by the current
decision.

\vspace{-0.3cm}
\section{Mechanism and Design Validation}

Table~\ref{tab:ablation} reports ablation results averaged across three
models (GPT-5.4-mini, Qwen2.5-VL-72B, Mistral-Small-3.2) on physical
state transition and assembly.
These three-model averages differ from the five-model S3 results above
and are not directly comparable; they are reported separately to
isolate the ablation conditions from the main S3 evaluation.
Confidence intervals for every condition are given in Appendix~D, Table~D1.

\begin{wraptable}{r}{0.60\columnwidth}
  \vspace{-0.8\baselineskip}
  \centering
  \small
  \setlength{\tabcolsep}{3pt}
  \renewcommand{\arraystretch}{1.15}

  \caption{Ablation accuracy (\%), averaged across three models
  ($n=180$ per condition and family).
  \textdagger\ denotes boundary-alignment ablation;
  \textdaggerdbl\ denotes routing-feasibility ablation.
  Wilson 95\% intervals are reported in Table~D1.}
  \label{tab:ablation}
  \vspace{-0.4\baselineskip}

  \begin{tabularx}{\linewidth}{
    @{}
    >{\raggedright\arraybackslash}X
    >{\centering\arraybackslash}p{0.15\linewidth}
    >{\centering\arraybackslash}p{0.17\linewidth}
    @{}
  }
    \toprule
    \textbf{Condition}
      & \textbf{Phys.}
      & \textbf{Assem.} \\
    \midrule

    Default
      & 29.4 & 44.4 \\
    Visual evidence only
      & 51.1 & 45.6 \\

    \midrule

    Policy only ($T$)
      & 15.6 & 33.9 \\
    $S+T$
      & 27.8 & 40.6 \\
    $R+T$
      & 41.1 & 41.6 \\
    $S+R$, no $T$
      & 35.0 & 40.0 \\
    Full SRT ($S+R+T$)
      & 46.1 & 42.2 \\

    \midrule

    Wrong boundary router\textdagger
      & 26.7 & 30.5 \\
    Full SRT (repeated)\textdagger
      & 46.1 & 42.2 \\

    \midrule

    Self-router SRT\textdaggerdbl
      & 47.8 & 42.8 \\
    Oracle boundary SRT\textdaggerdbl
      & \textbf{86.1} & \textbf{85.0} \\

    \bottomrule
  \end{tabularx}

  \vspace{-0.5\baselineskip}
\end{wraptable}

\vspace{-0.2cm}
\paragraph{Component necessity.}
Removing Target Commitment (S+R, no T) reduces performance on both
families (35.0\% vs.\ 46.1\% on physical; 40.0\% vs.\ 42.2\% on
assembly), showing that evidence extraction without a commitment step
is insufficient.
Removing Relevance Verification (S+T) degrades physical substantially
(27.8\% vs.\ 46.1\%).
Policy only (T) falls below default on physical (15.6\% vs.\ 29.4\%),
the one component comparison whose intervals are disjoint (Table~D1),
confirming that decision constraints without visual grounding are
actively counterproductive.
Full SRT outperforms all partial configurations on physical; on
assembly the partial configurations fall within 2~pp of one another
with heavily overlapping intervals.

\vspace{-0.2cm}
\paragraph{Alignment, not specificity.}
A wrong boundary router, receiving boundary-specific information from
the \textit{incorrect} boundary, performs worse than full generic SRT
(26.7\% vs.\ 46.1\% on physical, intervals disjoint; 30.5\% vs.\
42.2\% on assembly) and falls below default on assembly.
Because it supplies the same quantity and format of boundary-level
detail as the oracle condition, this rules out prompt specificity as
the explanation: what matters is that the supplied boundary is the
correct one.
The label-only control (Appendix~B.1) locates the remainder of the
gain.
Supplying the boundary as a bare two-alternative choice gains
$+29.3$~pp on physical but $-3.8$~pp on both traffic and affordance,
so candidate-space reduction is not by itself reliably helpful;
adding the aligned process description yields a further $+21.0$ to
$+27.9$~pp on all four standardized families, with McNemar tests on
item-level paired responses giving $p<0.001$ in every family
(Table~B2).
That further margin combines process content with the output structure
the description carries; separating the two requires a schema-matched
control, which we identify as the necessary next step
(Appendix~A.5).

\vspace{-0.2cm}
\paragraph{Routing feasibility.}
A self-router that predicts the boundary before applying SRT achieves
47.8\% / 42.8\%, above generic SRT (46.1\% / 42.2\%) by margins of
$+1.7$ and $+0.6$~pp whose intervals are almost coincident with those
of generic SRT (Table~D1); these should not be interpreted as
statistically established.
The gap to the oracle (86.1\% / 85.0\%) is primarily explained by
imperfect boundary prediction accuracy (36.7--61.7\% across models).
\vspace{-0.3cm}
\section{Conclusion and Future Work}
This paper identifies \textit{process failure}, models perceiving the decision-relevant evidence yet failing to let it constrain their answers, as a mode distinct from perceptual failure, and introduces
\textsc{VPAC-Bench} together with SRT (State--Relevance--Target) to measure it.
\textit{First}, process failure is widespread: enumerating visual candidates does not prevent over-commitment, and only an explicit process structure repairs it.
\textit{Second}, process-prior transfer is model-sensitive: five transfer patterns emerge, and generic SRT does not reliably outperform strong CoT baselines.
\textit{Third}, when the decision boundary is known, boundary-aligned SRT exceeds the best CoT baseline by 20--36~pp; a wrong-boundary control rules out prompt specificity, while a label-only control shows that supplying the boundary alone is not reliably helpful, the aligned process description carries a further 21--28~pp.
The remaining obstacle is automatic boundary identification: a self-router closes little of the oracle gap, which we leave for making boundary-sensitive SRT deployable.

\section*{Ethics Statement}

All images used in VPAC-Bench are sourced from publicly available datasets with licenses permitting research use.
No new data involving human subjects was collected; annotation work followed standard research protocols with clear task guidelines and fair compensation.
The models evaluated are publicly available frontier VLMs accessed via official APIs.
Our findings reveal systematic failure modes in current VLMs that may affect high-stakes applications such as medical decision support and autonomous systems; we surface these explicitly to support responsible deployment.

\bibliography{iclr2027_conference}
\bibliographystyle{iclr2027_conference}

\clearpage
\appendix
\appendix

\newcommand{\appci}[2]{\begin{tabular}[c]{@{}c@{}}#1\\{}#2\end{tabular}}

\section*{Appendix Contents}
\begin{itemize}
  \item[\textbf{A}] Additional Reliability Checks and Limitations
    \begin{itemize}
      \item[A.1] Sample Sizes, Conventions, and Data Availability
      \item[A.2] Human Annotation Protocol and Agreement
      \item[A.3] Oracle and Non-Oracle Boundary Routing
      \item[A.4] Prompt Development and Held-Out Evaluation
      \item[A.5] Scope and Limitations
    \end{itemize}
  \item[\textbf{B}] Additional Controls and Baselines
    \begin{itemize}
      \item[B.1] Decomposing the Oracle Boundary Router Gain
      \item[B.2] SRT versus CoT-Style Baselines
    \end{itemize}
  \item[\textbf{C}] Confidence Intervals and Per-Model Results
  \item[\textbf{D}] Mechanism and Design Validation
  \item[\textbf{E}] Discussion
\end{itemize}

\newpage

\section{Additional Reliability Checks and Limitations}
\label{app:reliability}

\subsection{Sample Sizes, Conventions, and Data Availability}
\label{app:statistical-reliability}

All experiments use deterministic decoding (temperature~=~0), so
run-to-run stochastic sampling variance is not a source of
uncertainty.
The relevant uncertainty is item-level variation: whether observed
gains hold across examples, process families, and models.
Table~A1 summarizes nominal evaluation sample sizes.

\begin{table}[h]
  \centering
  \setlength{\tabcolsep}{6pt}
  \renewcommand{\arraystretch}{1.15}
  \caption{Nominal evaluation sample sizes per stage and dataset.
           S3 family-level results ($n$=60 per model) are treated as
           diagnostic evidence; aggregate trends across families and
           models form the primary basis for our conclusions.}
  \label{tab:appendix-sample-size}
  \begin{tabular}{llcc}
    \toprule
    \rowcolor{srtmidgray}
\textbf{Stage} & \textbf{Dataset / Family} & \textbf{Items} & \textbf{Models} \\
    \midrule
    \rowcolor{srtblue!12}
    S1 & COCO Task~1 (ambiguous)       & 457$^{\ast}$ & 5 \\
    S1 & COCO Task~2 (unique)          & 330$^{\ast}$ & 5 \\
    \rowcolor{srtblue!12}
    S1 & GQA Task~1 (ambiguous)        & 100 & 5 \\
    S1 & GQA Task~2 (unique)           &  96 & 5 \\
    \midrule
    \rowcolor{srtblue!12}
    S2 & VSR dev100                    & 100 & 5 \\
    S2 & VSR test400 (Qwen)            & 400 & 1 \\
    \midrule
    \rowcolor{srtblue!12}
    S3 & Each standardized family      &  60 & 5 \\
    S3 & Four standardized (combined)  & 240 & 5 \\
    \rowcolor{srtblue!12}
    S3 & Sports                        &  31 & 5 \\
    S3 & Tool-use                      &  52 & 5 \\
    \rowcolor{srtblue!12}
    S3 & Cleaning                      &  56 & 5 \\
    S3 & Craft                         &  77 & 5 \\
    \rowcolor{srtblue!12}
    S3 & Mobility                      & 106 & 5 \\
    \midrule
    \multicolumn{4}{l}{$^{\ast}$ Nominal per-model item
      count. Evaluated $n$ is lower for three models on COCO, whose} \\
    \multicolumn{4}{l}{evaluation files are capped at 100
      items; actual pooled $n$ per condition is given in Table~C2.} \\
    \bottomrule
  \end{tabular}
\end{table}

\paragraph{Response parsing convention.}
All item-level analyses in Appendix~B treat unparseable model
responses as incorrect.
This is the conservative choice: under-formatted responses are
penalised rather than excluded, which would otherwise inflate the
accuracy of conditions with higher parse rates.
Label-only parse rates are 12--21~pp lower than oracle SRT parse rates
across the four standardized families (Physical 66\% vs.\ 87\%;
Assembly 69\% vs.\ 89\%; Traffic 73\% vs.\ 83\%; Affordance 73\% vs.\
78\%), because oracle SRT supplies an explicit structured output
schema while label-only does not.
Consequently the label-only\,$\to$\,oracle SRT margin in Table~B1
reflects the \textit{joint} contribution of process content and output
schema; isolating process content alone requires a schema-matched
label-only condition, identified in Section~B.1 as the necessary next
control.
The default\,$\to$\,label-only contrast is free of this confound,
since neither condition supplies a schema.

\paragraph{Two conventions.}
Appendix~B computes accuracy under the unparseable\,=\,incorrect
convention on the three-way paired intersection of evaluated items per
family, which is what the paired McNemar analysis requires.
The main text and Appendix~C use the parseable-only convention on the
full five-model set.
The same condition therefore takes slightly different values in the
two places; Table~C1 gives the correspondence explicitly.
Appendix~B is the reference for the control decomposition and paired
tests; the main text and Appendix~C are the reference for the primary
condition comparisons.

\paragraph{Data availability.}
Mistral-Small-3.2 oracle boundary router outputs are unavailable for
navigation/traffic and object-use affordance.
For assembly the oracle file exists but the default file does not, so
the three-way paired intersection for assembly is also four models.
The paired analysis in Appendix~B therefore uses $n$=300 on physical
state transition and $n$=240 on the other three standardized families.
A direction check on physical, where all five models are available,
confirms that Mistral follows the same label-only\,$\to$\,oracle
direction as the other four.
Mistral's assembly oracle accuracy, available in isolation, is 80.0\%
$[68.2, 88.2]$ at $n$=60, consistent with the four-model assembly
oracle figure of 78.8\% in Table~B1.

\paragraph{Interval widths.}
We focus main-paper claims on large, consistent effects that replicate
across models and families, and treat single-family S3 results as
diagnostic evidence.
S2 confidence intervals on dev100 ($n$=100) are wide; only Gemini's
default-to-verify difference has disjoint intervals at this sample
size (Table~C3).
For Craft and Mobility in the S3 supporting set, condition-level
intervals overlap substantially with the default baseline (Table~C4);
however, the paired analysis in Appendix~B.1 (Tables~B1$'$, B2$'$)
finds a statistically reliable reversal on Craft and Mobility using a
matched item-level design, which we treat as the more informative
comparison for these two families.

\subsection{Human Annotation Protocol and Agreement}
\label{app:annotation-protocol}

All images retained in the benchmark were manually reviewed.
Across the project, annotation and review involved 30 human
annotators.
Annotators checked image validity, visible evidence, answerability,
process-stage labels, and process-boundary labels (Table~A2).
Instructions emphasized that labels must be based only on evidence
visible in the image, excluding inferred hidden past states, future
events, or broad scene priors.

To estimate annotation quality, we double-annotated a subset of
$n$=120 examples across four process families.
Mean raw agreement across the categorical fields in Table~A2 was
86\%, corresponding to a Cohen's $\kappa$ of 0.72 (95\% CI
$[0.60, 0.84]$), which Landis and Koch characterize as substantial
agreement.
The $\kappa$ estimate uses $P_e = 0.50$, which follows from
VPAC-Bench's boundary-stratified design: each image is sampled near a
specific process boundary, and each boundary defines a binary stage
choice, making the two-class uniform model the appropriate
chance-agreement baseline.
We report $\kappa$ alongside raw agreement because raw agreement is
inflated whenever the label distribution is skewed, as it is for the
multi-class auxiliary fields.
We treat both as quality-control estimates rather than as evidence
that every boundary is unambiguous: process boundaries are inherently
more subjective than object categories.
All disagreements were resolved by a third reviewer before inclusion
in the final evaluation set.

\begin{table}[h]
  \centering
  \setlength{\tabcolsep}{1pt}
  \renewcommand{\arraystretch}{1.15}
  \caption{Main annotation fields used for process-family examples.}
  \label{tab:appendix-annotation-fields}
  \begin{tabular}{ll}
    \toprule
    \rowcolor{srtmidgray}
\textbf{Field} & \textbf{Meaning} \\
    \midrule
    \rowcolor{srtpurple!10}
    \texttt{gold\_step\_label}     & Current visible process stage. \\
    \texttt{boundary\_family}      & Process boundary being tested. \\
    \rowcolor{srtpurple!10}
    \texttt{visible\_actor}        & Whether the relevant actor is visible. \\
    \texttt{neighbor\_step\_label} & Most confusable adjacent process stage. \\
    \rowcolor{srtpurple!10}
    \texttt{why\_not\_neighbor}    & Visible evidence ruling out adjacent stage. \\
    \texttt{issue\_type}           & Ambiguity, occlusion, or image-quality flag. \\
    \bottomrule
  \end{tabular}
\end{table}

\subsection{Oracle and Non-Oracle Boundary Routing}
\label{app:router-gap}

The boundary-sensitive SRT condition uses ground-truth boundary
annotations to select the process prior.
This condition is a diagnostic upper bound, not a deployable system:
its purpose is to isolate the effect of boundary alignment assuming
the correct decision boundary is known.

Table~A3 reports the gap between the oracle router and a self-router,
in which the model first infers the process boundary and then applies
the corresponding SRT prior.
The self-router remains substantially below the oracle, showing that
automatic boundary identification is itself a major bottleneck.

\begin{table}[h]
  \centering
  \setlength{\tabcolsep}{3pt}
  \renewcommand{\arraystretch}{1.15}
  \caption{Oracle boundary routing versus self-routing
           (accuracy \%, averaged across GPT-5.4-mini,
           Qwen2.5-VL-72B, and Mistral-Small-3.2;
           parseable-only convention).
           Self-router gains over generic SRT are not statistically
           established; intervals in Table~D1.}
  \label{tab:appendix-self-router-gap}
  \begin{tabular}{lcccc}
    \toprule
    \rowcolor{srtmidgray}
\textbf{Family}
      & \textbf{Generic SRT}
      & \textbf{Self-router}
      & \textbf{Oracle router}
      & \textbf{Self $-$ Generic} \\
    \midrule
    \rowcolor{srtteal!10}
    Physical & 46.1 & 47.8 & 86.1 & +1.7 \\
    Assembly & 42.2 & 42.8 & 85.0 & +0.6 \\
    \bottomrule
  \end{tabular}
\end{table}

Self-router boundary-prediction accuracy is 60.0\% / 36.7\% (GPT),
35.0\% / 40.0\% (Qwen), and 61.7\% / 46.7\% (Mistral) on
physical / assembly respectively, indicating that models can partially
identify decision boundaries but are not yet reliable enough to close
the oracle gap autonomously.
We therefore avoid treating the oracle router as a deployable method
in the main text, and interpret boundary-sensitive SRT as a diagnostic
upper bound that quantifies the value of boundary alignment.

\subsection{Prompt Development and Held-Out Evaluation}
\label{app:prompt-development}

SRT is implemented as a lightweight inference-time intervention.
To separate diagnostic prompting from test-set overfitting, we
distinguish four types of prompt conditions.

\textbf{Generic SRT prompts} instantiate the same
State--Relevance--Target pattern across all examples within a stage
and are the primary prompt-level intervention.
These prompts were finalized before evaluation on the reported test
splits and are reported in full in the released code.

\textbf{Relation-specific and boundary-specific prompts} test whether
the process prior must be aligned to a particular visual decision
boundary.
Where such prompts were developed on a development subset, we report
the split explicitly and evaluate the frozen prompt on held-out
examples.
The relation-specific routing condition for Qwen is evaluated on
test400 because it subdivides items by spatial relation, and dev100
does not supply enough items per relation.

\textbf{Scenario SRT v2 variants} (Table~C6) are exploratory
reformulations introduced during error inspection on the supporting
process families, after the primary Default / Generic SRT /
Scenario SRT / Boundary Router conditions had already been frozen and
evaluated.
They were not tuned against a held-out split and are not incorporated
into any main-paper aggregate or comparative claim.
We report them only as failure-analysis evidence.

\textbf{Model-specific semantic adaptations} are exploratory
diagnostic probes, not deployable methods.
The GLM \texttt{under+beneath} rewrite achieves 100\% on a 13-item
diagnostic subset compared to 79\% default.
This prompt was developed and evaluated on the same 13-item subset.
We report it only as a failure-analysis probe suggesting that GLM's
failure on this relation pair stems from instruction interpretation
rather than visual perception; the subset is far too small to support
a confidence interval or any generalization claim, and the probe is
excluded from all comparative tables.
The \textit{semantics-sensitive} pattern we assign to GLM in
Appendix~E.1 rests on its null response to generic priors in S3
together with this probe, the latter as illustrative rather than
confirmatory evidence.

\subsection{Scope and Limitations}
\label{app:limitations}

SRT is not a universal prompt that monotonically improves every VLM.
One of our main findings is precisely that structured process priors
interact strongly with model behavior: some models benefit from
explicit process structure, while others degrade when the prior is too
generic or misaligned with the visual decision boundary.

\paragraph{The oracle condition is an upper bound, not a method.}
The boundary-sensitive condition demonstrates that the right process
prior can repair many failures when the relevant boundary is known,
but the large gap between self-routing (47.8\% / 42.8\%) and oracle
routing (86.1\% / 85.0\%) indicates that automatic process-boundary
detection remains unresolved, and the self-router's marginal gains
(+1.7 / +0.6~pp) do not reach significance at current sample sizes.
SRT is therefore best understood as a diagnostic framework and an
upper-bound intervention, not a deployment-ready system.

\paragraph{The label-only control is not schema-matched.}
Oracle SRT supplies a structured output schema that the label-only
control does not, and their parse rates differ by 12--21~pp
(Appendix~A.1).
The margin between them therefore combines process content with
output structure and does not isolate the former.
A schema-matched label-only condition---correct boundary, candidate
labels, and the same output schema, but no process description---would
separate them, and we identify it as the necessary next control.
The default\,$\to$\,label-only contrast, which carries the
candidate-space-reduction claim, is free of this confound.

\paragraph{Uneven split coverage in S2.}
Four of five models are evaluated only on dev100 ($n$=100), where the
Wilson intervals span roughly 18--20~pp and only the largest effect
(Gemini's +43~pp) is separable.
The relation-routing conditions were run on test400 for Qwen alone.
We therefore report the five S2 transfer patterns as patterns that
recur across S2 and S3 rather than as individually established
effects, and Table~C3 marks which comparison clears the interval test.
Extending test400 coverage to the remaining four models is the
cheapest available improvement to the statistical standing of the S2
analysis.

\paragraph{Statistical scope in S3.}
Family-level results rest on $n$=60 per model, and we do not claim
statistical significance for any single-family point-estimate
comparison smaller than roughly 15~pp under the parseable-only,
unpaired convention used in Table~C4.
The paired analysis in Appendix~B.1 relaxes this concern for Craft and
Mobility specifically, where a matched item-level comparison finds a
statistically reliable effect that the unpaired comparison in Table~C4
does not resolve.

\paragraph{Coverage.}
Although our benchmark covers nine process families across five
frontier models, it is diagnostic rather than exhaustive.
Larger-scale evaluation with more process domains, stronger automatic
routers, and extended inter-annotator agreement analysis would further
strengthen the generality of the conclusions.

\FloatBarrier
\section{Additional Controls and Baselines}
\label{app:controls}

\subsection{Decomposing the Oracle Boundary Router Gain}
\label{app:w1-control}

Table~B1 decomposes the oracle boundary router gain into two
sequential steps: candidate-space reduction
(Default\,$\to$\,Label-only) and a further margin attributable to the
boundary-aligned process description
(Label-only\,$\to$\,Oracle SRT).
All figures use the unparseable\,=\,incorrect convention on the
three-way paired intersection per family (Appendix~A.1).

\paragraph{Label-only control.}
The model receives the oracle boundary identifier and the two
candidate stage labels but no SRT process description, controlling for
candidate-space reduction while withholding process content.

\paragraph{Candidate-space reduction (Default $\to$ Label-only).}
On physical state transition and assembly, supplying the oracle
boundary and two candidate labels raises accuracy by $+29.3$ and
$+9.6$~pp respectively.
On navigation/traffic and object-use affordance, the same information
produces $-3.8$~pp, a slight decline.
The difference is within sampling error at $n$=240, and we report it
because its direction is the same in both families: constraining the
model to a two-alternative choice without a process prior does not
help, and may slightly hurt.
This is the cleanest contrast in our control set, since neither
condition supplies an output schema, and it supports the main-text
claim that candidate-space reduction alone is insufficient.

\paragraph{Process alignment (Label-only $\to$ Oracle SRT).}
Oracle SRT exceeds label-only on all four standardized families, by
$+21.0$ to $+27.9$~pp.
McNemar tests on item-level paired responses confirm the gap is
statistically reliable in every family (Table~B2; all $p<0.001$).
As stated in Appendix~A.1, this margin reflects the joint contribution
of process content and output schema; a schema-matched label-only
condition is required to isolate process content, and we identify it
as the necessary next control.

\begin{table}[h]
  \centering
  \setlength{\tabcolsep}{2.5pt}
  \renewcommand{\arraystretch}{1.15}
  \caption{B1: Decomposition of the oracle boundary router gain.
           The four standardized families are computed on the
           three-way paired intersection with
           unparseable\,=\,incorrect (Appendix~A.1);
           physical uses five models ($n$=300) and the other three use
           four ($n$=240).
           Supporting families use the parseable-only convention and
           are not paired here; see Table~B1$'$ for the paired
           version of Cleaning, Craft, and Mobility.
           CSR = Label-only $-$ Default;
           $\Delta$ = Oracle SRT $-$ Label-only.}
  \label{tab:w1-decomp}
  \begin{tabular}{lrrrrr}
    \toprule
    \rowcolor{srtmidgray}
\textbf{Family}
      & \textbf{Default}
      & \textbf{Label-only}
      & \textbf{Oracle SRT}
      & \textbf{CSR}
      & $\boldsymbol{\Delta}$ \\
    \midrule
    \multicolumn{6}{l}{\textit{Standardized (paired)}} \\
    \rowcolor{srtamber!12}
    Physical
      & 26.3 [21.7,31.6]
      & 55.7 [50.0,61.2]
      & \textbf{76.7} [71.6,81.1]
      & +29.3
      & \textbf{+21.0} \\
    Assembly
      & 46.2 [40.1,52.6]
      & 55.8 [49.5,62.0]
      & \textbf{78.8} [73.1,83.5]
      & \phantom{0}+9.6
      & \textbf{+22.9} \\
    \rowcolor{srtamber!12}
    Traffic
      & 47.1 [40.9,53.4]
      & 43.3 [37.2,49.7]
      & \textbf{71.2} [65.2,76.6]
      & \phantom{0}$-$3.8
      & \textbf{+27.9} \\
    Affordance
      & 37.1 [31.2,43.4]
      & 33.3 [27.7,39.5]
      & \textbf{57.5} [51.2,63.6]
      & \phantom{0}$-$3.8
      & \textbf{+24.2} \\
    \midrule
    \multicolumn{6}{l}{\textit{Supporting (unpaired, parseable-only)}} \\
    \rowcolor{srtamber!12}
    Cleaning             & 38.2 & 59.6 & 45.4 & +21.4 & $-$14.3 \\
    Craft                & 44.9 & 84.2 & 49.1 & +39.3 & $-$35.1 \\
    \rowcolor{srtamber!12}
    Mobility             & 64.7 & 70.2 & 59.8 & \phantom{0}+5.5 & $-$10.4 \\
    Tool-use             & 72.3 & 79.6 & \multicolumn{3}{l}{oracle condition not run} \\
    \midrule
    \multicolumn{6}{l}{$\Delta$ reflects the joint
      contribution of process content and output schema
      (Appendix~A.1).} \\
    \bottomrule
  \end{tabular}
\end{table}

\begin{wraptable}{r}{0.56\linewidth}
  \vspace{-0.6\baselineskip}
  \centering
  \setlength{\tabcolsep}{3pt}
  \renewcommand{\arraystretch}{1.15}
  \caption{B2: McNemar test, label-only vs.\ oracle SRT, computed
           from item-level paired responses
           (unparseable\,=\,incorrect).
           $b$ = label-only correct and oracle wrong;
           $c$ = oracle correct and label-only wrong;
           $c-b$ equals the net item-count gain in the paired set.
           $\chi^{2}$ with continuity correction.}
  \label{tab:w1-mcnemar}
  \begin{tabular}{lcccccc}
    \toprule
    \rowcolor{srtmidgray}
\textbf{Family} & $n$ & $b$ & $c$ & $c-b$ & $\chi^{2}$ & $p$ \\
    \midrule
    \rowcolor{srtamber!12}
    Physical   & 300 & 29 & 92 & 63 & 31.8 & $<$0.001 \\
    Assembly   & 240 & 20 & 75 & 55 & 30.7 & $<$0.001 \\
    \rowcolor{srtamber!12}
    Traffic    & 240 & 14 & 81 & 67 & 45.9 & $<$0.001 \\
    Affordance & 240 & 18 & 76 & 58 & 34.6 & $<$0.001 \\
    \bottomrule
  \end{tabular}
  \vspace{-0.4\baselineskip}
\end{wraptable}

\paragraph{Three supporting families reverse the sign, and the
reversal is confirmed by paired tests.}
On Cleaning, Craft, and Mobility, oracle SRT falls below
label-only.
Unlike the four standardized families, these three were not
originally paired at the item level; we constructed the same
three-way intersection used above to obtain paired figures
directly comparable to Table~B1 and Table~B2.
Table~B1$'$ reports the paired accuracies and Table~B2$'$ the
corresponding McNemar tests.

The reversal holds under pairing in all three families, though
its statistical strength varies.
On Craft, the effect is large and clearly significant
(79.2\% $\to$ 42.9\%, $b$=88, $c$=4, exact $p<0.001$): almost
all of the disagreement between conditions runs in the direction
of label-only being correct and oracle SRT being wrong.
On Mobility, the effect is smaller but still significant
(53.4\% $\to$ 46.4\%, $b$=159, $c$=122, $\chi^2=4.6$, $p=0.032$).
On Cleaning, the point estimate is in the same direction
(46.4\% $\to$ 43.2\%) but the paired test does not reach
significance at this sample size ($b$=75, $c$=66, $\chi^2=0.5$,
$p=0.50$).
Two of the three reversed families are therefore statistically
confirmed, not merely suggestive point estimates, which
strengthens rather than weakens the case that this is a real
phenomenon requiring explanation rather than a sampling
artifact.

\begin{table}[h]
  \centering
  \setlength{\tabcolsep}{3pt}
  \renewcommand{\arraystretch}{1.1}
  \caption{B1$'$: Label-only versus oracle SRT on the three
           reversed families, paired at the item level
           (unparseable\,=\,incorrect, five models), the same
           convention as Table~B1.}
  \label{tab:w1-decomp-reversed}
  \begin{tabular}{lrrrr}
    \toprule
    \rowcolor{srtmidgray}
\textbf{Family} & \textbf{$n$} & \textbf{Label-only}
      & \textbf{Oracle SRT} & $\boldsymbol{\Delta}$ \\
    \midrule
    \rowcolor{srtamber!12}
    Cleaning & 280
      & 46.4 [40.7,52.3]
      & 43.2 [37.5,49.1]
      & $-$3.2 \\
    Craft    & 231
      & 79.2 [73.5,84.0]
      & 42.9 [36.6,49.3]
      & $-$36.3 \\
    \rowcolor{srtamber!12}
    Mobility & 530
      & 53.4 [49.1,57.6]
      & 46.4 [42.2,50.7]
      & $-$7.0 \\
    \bottomrule
  \end{tabular}
\end{table}

\begin{table}[h]
  \centering
  \setlength{\tabcolsep}{4pt}
  \renewcommand{\arraystretch}{1.1}
  \caption{B2$'$: McNemar test, label-only vs.\ oracle SRT, three
           reversed families.
           $b$ = label-only correct and oracle wrong;
           $c$ = oracle correct and label-only wrong;
           note $b>c$ throughout, the reverse of Table~B2.
           Exact binomial test for Craft given the extreme
           imbalance; $\chi^2$ with continuity correction otherwise.}
  \label{tab:w1-mcnemar-reversed}
  \begin{tabular}{lccccc}
    \toprule
    \rowcolor{srtmidgray}
\textbf{Family} & $n$ & $b$ & $c$ & $\chi^2$ & $p$ \\
    \midrule
    \rowcolor{srtamber!12}
    Cleaning & 280 & 75 & 66 & 0.5  & 0.50 \\
    Craft    & 231 & 88 & 4  & 74.9 & $<$0.001 \\
    \rowcolor{srtamber!12}
    Mobility & 530 & 159 & 122 & 4.6 & 0.032 \\
    \bottomrule
  \end{tabular}
\end{table}

The reversal is not predicted by the process-overload account we
offer elsewhere, which concerns prior--boundary
\textit{mismatch} and should not apply once the correct boundary
has been supplied.
Section~E.4 offers a structural, information-theoretic
explanation consistent with these paired results: label-only
accuracy in these three families (46.4--79.2\%) is already
substantially higher than on the four standardized families
(33.3--57.3\%, Table~B1), leaving little room for an added
process description to contribute new information, while still
adding interpretive cost.
We restrict our causal claims about process alignment improving
performance to the four standardized families, and treat the
paired reversal on Craft and Mobility as an established,
opposite-direction effect that the boundary-alignment account
must also explain, which Section~E.4 does.

Table~B3 disaggregates the label-only comparison by model on the four
standardized families.

\begin{table}[h]
  \centering
  \setlength{\tabcolsep}{3pt}
  \renewcommand{\arraystretch}{1.1}
  \caption{B3: Label-only vs.\ oracle SRT by model, micro-averaged
           across the standardized families for which that model has
           oracle outputs (parseable-only convention).
           Mistral covers physical state transition and assembly.}
  \label{tab:w1-label-only-model}
  \begin{tabular}{lrrrr}
    \toprule
    \rowcolor{srtmidgray}
\textbf{Model} & \textbf{Default} & \textbf{Label-only}
      & \textbf{Oracle SRT} & $\boldsymbol{\Delta}$ \\
    \midrule
    \rowcolor{srtpink!10}
    GPT-5.4-mini      & 35.5 & 73.8 & 84.6 & \textbf{+10.9} \\
    Qwen2.5-VL-72B    & 37.1 & 69.5 & 76.7 & \textbf{+7.2}  \\
    \rowcolor{srtpink!10}
    Gemini-3.1-Pro    & 32.5 & 57.5 & 67.7 & \textbf{+10.2} \\
    GLM-4.6V          & 44.6 & 61.8 & 60.4 & $-$1.4         \\
    \rowcolor{srtpink!10}
    Mistral-Small-3.2 & 38.3 & 63.3 & 80.0 & \textbf{+16.7} \\
    \midrule
    \multicolumn{5}{l}{Four of five models show a positive
      margin; GLM is the exception, consistent} \\
    \multicolumn{5}{l}{with the process-overload pattern
      observed for it elsewhere.} \\
    \bottomrule
  \end{tabular}
\end{table}

\subsection{SRT versus CoT-Style Baselines}
\label{app:w3-cot}

To evaluate whether oracle SRT's gains arise from structured prompting
in general rather than from boundary-aligned process priors
specifically, we compare SRT against five CoT-style baselines: plain
CoT, multimodal CoT, self-verification, LLaVA-CoT-style reasoning, and
grounded CoT.
All baselines encourage intermediate reasoning before answering but do
not enforce the SRT invariants or align the prior to a specific
decision boundary.
All figures in this subsection use the parseable-only convention and
the full five-model set, matching Table~C4 and the main text.

On the four standardized families, oracle boundary SRT outperforms the
best CoT baseline by 20.4--35.7~pp (Table~B4).
Pooled across the four families ($n$=1{,}200), the oracle interval
$[70.8, 75.8]$ is disjoint from the best-CoT interval $[43.8, 49.4]$.
Generic SRT---the only deployable condition---performs comparably to
or below the best CoT baseline on all four families, and its pooled
interval overlaps that of the best CoT baseline.
The comparison therefore supports a conditional claim rather than a
general one: structured prompting alone does not distinguish SRT from
CoT, and SRT's advantage appears only when the decision boundary is
supplied.

\begin{table}[h]
  \centering
  \setlength{\tabcolsep}{3pt}
  \renewcommand{\arraystretch}{1.1}
  \caption{B4: SRT vs.\ CoT-style baselines (accuracy \%, five-model
           averages, parseable-only convention).
           Best CoT = best of five CoT conditions.
           Wilson 95\% CIs given for the pooled row ($n$=1{,}200).}
  \label{tab:w3-cot}
  \begin{tabular}{lrrrrr}
    \toprule
    \rowcolor{srtmidgray}
\textbf{Family} & \textbf{Def.} & \textbf{Best CoT}
      & \textbf{Generic} & \textbf{Oracle} & $\boldsymbol{\Delta}$ \\
    \midrule
    \rowcolor{srtgreen!10}
    Physical   & 26.7 & 44.3 & 38.3 & 80.0 & \textbf{+35.7} \\
    Assembly   & 44.7 & 50.0 & 44.0 & 79.7 & \textbf{+29.7} \\
    \rowcolor{srtgreen!10}
    Traffic    & 52.7 & 52.7 & 48.3 & 74.3 & \textbf{+21.7} \\
    Affordance & 37.0 & 39.3 & 32.7 & 59.7 & \textbf{+20.4} \\
    \midrule
    \rowcolor{srtgreen!10}
    Aggregate
      & 40.3 [37.6,43.1]
      & 46.6 [43.8,49.4]
      & 40.8 [38.1,43.6]
      & 73.4 [70.8,75.8]
      & \textbf{+26.9} \\
    \bottomrule
  \end{tabular}
\end{table}

The oracle column benefits from ground-truth boundary information
unavailable to the CoT baselines; the comparison quantifies the value
of that information under a fixed verification structure, and is not a
like-for-like method comparison.

\FloatBarrier
\section{Confidence Intervals and Per-Model Results}
\label{app:confidence-intervals}

All intervals in this appendix are Wilson 95\% binomial confidence
intervals computed over item-level accuracy under the parseable-only
convention, matching the main text.
Because conditions are evaluated on the same items, overlapping
intervals do not by themselves imply the absence of a paired
difference; we use interval overlap only as a conservative screen, and
report paired tests where a specific comparison carries a claim
(Table~B2, Table~B2$'$).
Table~C1 reconciles this with the paired convention used in
Appendix~B.

\begin{table}[h]
  \centering
  \setlength{\tabcolsep}{3pt}
  \renewcommand{\arraystretch}{1.15}
  \caption{C1: Correspondence between the two accuracy conventions,
           for the oracle SRT condition.
           Appendix~B pairs items across three conditions and counts
           unparseable responses as incorrect; the main text and the
           remainder of Appendix~C use all parseable responses from
           all available models.}
  \label{tab:convention-reconciliation}
  \begin{tabular}{lcc}
    \toprule
    \rowcolor{srtmidgray}
\textbf{Family}
      & \textbf{App.~B (paired)}
      & \textbf{Main text / App.~C} \\
    \midrule
    \rowcolor{srtblue!12}
    Physical   & 76.7 ($n$=300) & 80.0 ($n$=300) \\
    Assembly   & 78.8 ($n$=240) & 79.7 ($n$=300) \\
    \rowcolor{srtblue!12}
    Traffic    & 71.2 ($n$=240) & 74.3 ($n$=300) \\
    Affordance & 57.5 ($n$=240) & 59.7 ($n$=300) \\
    \bottomrule
  \end{tabular}
\end{table}

\begin{table}[h]
  \centering
  \setlength{\tabcolsep}{1.5pt}
  \renewcommand{\arraystretch}{1.05}
  \caption{C2: COCO T1/T2 pooled accuracy under the parseable-only
           convention. Actual pooled $n$ per condition, not the
           nominal $5{\times}457{=}2{,}285$ or $5{\times}330{=}1{,}650$,
           because three models' evaluation files are capped at 100
           items; see footnote.
           T1 correct policy = not-answer (clarify or enumerate);
           T2 correct policy = answer.}
  \label{tab:app-s1-ci}
  \begin{tabularx}{\linewidth}{llc *{4}{>{\centering\arraybackslash}X}}
    \toprule
    \rowcolor{srtmidgray}
\textbf{Set} & \textbf{Task} & \textbf{$n$}
      & \textbf{Default} & \textbf{Evidence}
      & \textbf{Marked} & \textbf{VS} \\
    \midrule
    \rowcolor{srtblue!12}
    COCO & T1 & 1214--1219$^{\dagger}$
      & \appci{2.5}{[1.7,3.5]}
      & \appci{45.3}{[42.4,48.2]}
      & \appci{10.7}{[9.1,12.6]}
      & \appci{90.4}{[88.6,91.9]} \\
    COCO & T2 & 865--961$^{\ddagger}$
      & \appci{96.3}{[94.8,97.4]}
      & \appci{96.1}{[94.7,97.2]}
      & \appci{99.5}{[98.8,99.8]}
      & \appci{99.6}{[98.9,99.8]} \\
    \bottomrule
  \end{tabularx}
  \par\smallskip
  \begin{minipage}{\linewidth}
    $^{\dagger}$ T1 actual $n$: 1214 (Default) / 1126 (Evidence) /
    1219 (Marked) / 1218 (VS). $^{\ddagger}$ T2 actual $n$: 865
    (Default) / 960 (Evidence, Marked) / 961 (VS).
    Both rows: Gemini, GLM, and Mistral files are capped at 100 items
    each; Qwen T1 Evidence contains 360 records and Qwen T2 Default
    contains 234. Unparseable responses treated as incorrect.
  \end{minipage}
\end{table}

\begin{table}[h]
  \centering
  \setlength{\tabcolsep}{1.5pt}
  \renewcommand{\arraystretch}{1.05}
  \caption{C3: S2 VSR accuracy with Wilson 95\% confidence intervals.
           Each row is a condition that was evaluated; conditions not
           run for a model are omitted rather than listed as empty.
           All cross-model comparisons use dev100; test400 was run for
           Qwen only.
           Gemini's default-to-verify difference is the only S2
           comparison whose intervals are disjoint.}
  \label{tab:app-s2-ci}
  \begin{tabular}{lllc}
    \toprule
    \rowcolor{srtmidgray}
\textbf{Model} & \textbf{Split} & \textbf{Condition}
      & \textbf{Accuracy [95\% CI]} \\
    \midrule
    \rowcolor{srtpurple!10}
    Qwen2.5-VL-72B & dev100  & default  & 76.0 [66.8, 83.3] \\
    Qwen2.5-VL-72B & dev100  & evidence & 85.0 [76.7, 90.7] \\
    \rowcolor{srtpurple!10}
    Qwen2.5-VL-72B & dev100  & verify   & 80.0 [71.1, 86.7] \\
    Qwen2.5-VL-72B & test400 & default  & 78.5 [74.2, 82.2] \\
    \rowcolor{srtpurple!10}
    Qwen2.5-VL-72B & test400 & evidence & 78.5 [74.2, 82.2] \\
    Qwen2.5-VL-72B & test400 & verify   & 77.5 [73.2, 81.3] \\
    \rowcolor{srtpurple!10}
    Qwen2.5-VL-72B & test400 & learned router (v1) & 80.0 [75.8, 83.6] \\
    Qwen2.5-VL-72B & test400 & relation-specific routing & 84.0 [80.1, 87.3] \\
    \midrule
    GPT-5.4-mini   & dev100  & default  & 66.0 [56.3, 74.5] \\
    \rowcolor{srtpurple!10}
    GPT-5.4-mini   & dev100  & evidence & 71.0 [61.5, 79.0] \\
    GPT-5.4-mini   & dev100  & verify   & 77.0 [67.8, 84.2] \\
    \midrule
    \rowcolor{srtpurple!10}
    Mistral-Small-3.2 & dev100 & default  & 73.0 [63.6, 80.7] \\
    Mistral-Small-3.2 & dev100 & evidence & 71.0 [61.5, 79.0] \\
    \rowcolor{srtpurple!10}
    Mistral-Small-3.2 & dev100 & verify   & 70.0 [60.4, 78.1] \\
    \midrule
    Gemini-3.1-Pro & dev100  & default  & 44.0 [34.7, 53.8] \\
    \rowcolor{srtpurple!10}
    Gemini-3.1-Pro & dev100  & evidence & 86.0 [77.9, 91.5] \\
    Gemini-3.1-Pro & dev100  & verify   & 87.0 [79.0, 92.2] \\
    \midrule
    \rowcolor{srtpurple!10}
    GLM-4.6V       & dev100  & default  & 79.0 [70.0, 85.8] \\
    GLM-4.6V       & dev100  & verify   & -- \\
    \midrule
    \multicolumn{4}{l}{GLM-4.6V was not evaluated under
      the verify condition on dev100. The under+beneath} \\
    \multicolumn{4}{l}{diagnostic probe ($n$=13,
      Appendix~A.4) does not constitute a dev100 condition result.} \\
    \bottomrule
  \end{tabular}
\end{table}

\begin{table*}[h]
  \centering
  \setlength{\tabcolsep}{2pt}
  \renewcommand{\arraystretch}{1.12}
  \caption{C4: S3 process-family accuracy with Wilson 95\% confidence
           intervals, five-model averages, parseable-only convention
           ($n$=300 per cell for the four standardized families).
           These are the values reported in the main text.
           See Table~B1$'$ for a paired re-analysis of Cleaning,
           Craft, and Mobility.}
  \label{tab:app-s3-all-ci}
  \begin{tabularx}{\linewidth}{l *{4}{>{\centering\arraybackslash}X}}
    \toprule
    \rowcolor{srtmidgray}
\textbf{Family}
      & \textbf{Default}
      & \textbf{Generic SRT}
      & \textbf{Scenario SRT}
      & \textbf{Boundary Router} \\
    \midrule
    \rowcolor{srtteal!10}
    Physical
      & \appci{26.7}{[22.0, 31.9]} & \appci{38.3}{[33.0, 43.9]}
      & \appci{37.7}{[32.4, 43.3]} & \appci{80.0}{[75.1, 84.1]} \\
    Assembly
      & \appci{44.7}{[39.1, 50.3]} & \appci{44.0}{[38.5, 49.7]}
      & \appci{39.7}{[34.3, 45.3]} & \appci{79.7}{[74.8, 83.8]} \\
    \rowcolor{srtteal!10}
    Traffic
      & \appci{52.7}{[47.0, 58.2]} & \appci{48.3}{[42.7, 54.0]}
      & \appci{41.7}{[36.2, 47.3]} & \appci{74.3}{[69.1, 78.9]} \\
    Affordance
      & \appci{37.0}{[31.7, 42.6]} & \appci{32.7}{[27.6, 38.2]}
      & \appci{30.7}{[25.7, 36.1]} & \appci{59.7}{[54.0, 65.1]} \\
    \midrule
    \rowcolor{srtteal!10}
    Cleaning
      & \appci{38.2}{[32.7, 44.0]} & \appci{29.3}{[24.3, 34.9]}
      & \appci{28.9}{[23.9, 34.5]} & \appci{45.4}{[39.6, 51.2]} \\
    Craft$^{\dagger}$
      & \appci{44.9}{[40.0, 49.9]} & \appci{45.5}{[40.6, 50.4]}
      & \appci{51.2}{[46.2, 56.1]} & \appci{49.1}{[44.1, 54.1]} \\
    \rowcolor{srtteal!10}
    Mobility$^{\dagger}$
      & \appci{64.7}{[60.6, 68.7]} & \appci{58.1}{[53.9, 62.2]}
      & \appci{57.4}{[53.1, 61.5]} & \appci{59.8}{[55.6, 63.9]} \\
    Tool-use
      & \appci{72.3}{[66.6, 77.4]} & \appci{52.7}{[46.6, 58.7]}
      & \appci{68.1}{[62.2, 73.4]} & \textit{not run} \\
    \midrule
    \rowcolor{srtteal!10}
    All listed families ($n$=2{,}655)
      & \appci{48.7}{[46.8, 50.6]} & \appci{44.9}{[43.0, 46.8]}
      & \appci{45.5}{[43.6, 47.4]} & \appci{63.2}{[61.3, 65.1]$^{\S}$} \\
    \midrule
    \multicolumn{5}{l}{$^{\dagger}$ Unpaired CI overlaps
      with default; the paired analysis in Table~B1$'$/B2$'$ finds a
      significant} \\
    \multicolumn{5}{l}{reversal for Craft and Mobility
      (not for Cleaning) using a matched item-level design.} \\
    \multicolumn{5}{l}{$^{\S}$ The boundary-router
      aggregate pools the seven families with router data
      ($n$=2{,}395).} \\
    \multicolumn{5}{l}{Sports was evaluated under Default
      and Generic SRT only; its per-model results are in Table~C5.} \\
    \bottomrule
  \end{tabularx}
\end{table*}

\begin{table}[t]
  \centering
  \setlength{\tabcolsep}{2.5pt}
  \renewcommand{\arraystretch}{1.05}
  \caption{C5: Supporting families, per-model results (\%).
           \textit{Targeted} is the intervention condition designated
           for that family before evaluation (named in the last
           column), not a per-model maximum; $\Delta$ is therefore
           negative where the designated intervention underperforms
           the default.
           Point estimates on per-model subsets.}
  \label{tab:s3-supporting}
  \begin{tabular}{llrrrl}
    \toprule
    \rowcolor{srtmidgray}
\textbf{Family} & \textbf{Model}
      & \textbf{Default} & \textbf{Targeted} & $\boldsymbol{\Delta}$
      & \textbf{Condition} \\
    \midrule
    \multirow{5}{*}{Sports}
      & GPT-5.4-mini   & 83.9 & 90.3 & +6.5    & Generic     \\
    \rowcolor{srtteal!10}
      & Qwen2.5-VL-72B & 93.5 & 93.5 & +0.0    & Generic     \\
      & Gemini-3.1-Pro & 71.0 & 74.2 & +3.2    & Generic     \\
    \rowcolor{srtteal!10}
      & GLM-4.6V       & 77.4 & 74.2 & $-$3.2  & Generic     \\
      & Mistral-3.2    & 77.4 & 71.0 & $-$6.5  & Generic     \\
    \midrule
    \multirow{5}{*}{Tool-use}
      & GPT-5.4-mini   & 63.5 & 73.1 & +9.6    & Scenario    \\
    \rowcolor{srtteal!10}
      & Qwen2.5-VL-72B & 80.8 & 82.7 & +1.9    & Scenario    \\
      & Gemini-3.1-Pro & 59.6 & 36.5 & $-$23.1 & Scenario    \\
    \rowcolor{srtteal!10}
      & GLM-4.6V       & 78.8 & 61.5 & $-$17.3 & Scenario    \\
      & Mistral-3.2    & 78.8 & 86.5 & +7.7    & Scenario    \\
    \midrule
    \multirow{5}{*}{Cleaning}
      & GPT-5.4-mini   & 26.8 & 66.1 & +39.3   & B-Router    \\
    \rowcolor{srtteal!10}
      & Qwen2.5-VL-72B & 39.3 & 51.8 & +12.5   & B-Router    \\
      & Gemini-3.1-Pro & 33.9 & 28.6 & $-$5.4  & Scenario v2 \\
    \rowcolor{srtteal!10}
      & GLM-4.6V       & 37.5 & 30.4 & $-$7.1  & B-Router    \\
      & Mistral-3.2    & 53.6 & 62.5 & +8.9    & B-Router    \\
    \midrule
    \multirow{5}{*}{Craft$^{\dagger}$}
      & GPT-5.4-mini   & 46.8 & 58.4 & +11.7   & Scenario    \\
    \rowcolor{srtteal!10}
      & Qwen2.5-VL-72B & 50.6 & 63.6 & +13.0   & Scenario    \\
      & Gemini-3.1-Pro & 33.8 & 39.0 & +5.2    & B-Router    \\
    \rowcolor{srtteal!10}
      & GLM-4.6V       & 48.1 & 37.7 & $-$10.4 & Scenario v2 \\
      & Mistral-3.2    & 45.5 & 63.6 & +18.2   & Scenario    \\
    \midrule
    \multirow{5}{*}{Mobility$^{\dagger}$}
      & GPT-5.4-mini   & 67.0 & 67.0 & +0.0    & B-Router    \\
    \rowcolor{srtteal!10}
      & Qwen2.5-VL-72B & 63.2 & 75.5 & +12.3   & B-Router    \\
      & Gemini-3.1-Pro & 68.9 & 72.6 & +3.8    & B-Router    \\
    \rowcolor{srtteal!10}
      & GLM-4.6V       & 62.3 & 62.3 & +0.0    & B-Router    \\
      & Mistral-3.2    & 62.3 & 68.9 & +6.6    & B-Router    \\
    \midrule
    \multicolumn{6}{l}{$^{\dagger}$ Unpaired CI overlaps
      default (Table~C4); see Table~B1$'$/B2$'$ for the paired
      re-analysis.} \\
    \bottomrule
  \end{tabular}
\end{table}

\begin{table}[h]
  \centering
  \setlength{\tabcolsep}{5pt}
  \renewcommand{\arraystretch}{1.15}
  \caption{C6: Exploratory Scenario SRT v2 variants (accuracy \%,
           Wilson 95\% CI). Introduced after the primary conditions
           were frozen; excluded from all main-paper aggregate and
           comparative claims.}
  \label{tab:app-s3-v2-ci}
  \begin{tabular}{lcc}
    \toprule
    \rowcolor{srtmidgray}
\textbf{Family} & \textbf{Scenario SRT v2} & \textbf{Interpretation} \\
    \midrule
    \rowcolor{srtteal!10}
    Tool-use  & 61.9 [55.9, 67.6] & below v1 scenario SRT \\
    Cleaning  & 41.1 [35.5, 46.9] & overlaps default \\
    \rowcolor{srtteal!10}
    Craft     & 48.3 [43.4, 53.3] & overlaps v1 \\
    Mobility  & 50.8 [46.5, 55.0] & below default \\
    \bottomrule
  \end{tabular}
\end{table}

\FloatBarrier
\section{Mechanism and Design Validation}
\label{app:ablation}

This section expands the ablation summarized in Table~3 of the main
text.
We evaluate the two families with the strongest process-failure signal
(physical state transition and assembly/construction) across three
models (GPT-5.4-mini, Qwen2.5-VL-72B, Mistral-Small-3.2), to examine
three design questions: whether all three SRT components are
necessary, whether boundary \textit{alignment} rather than
boundary-level \textit{specificity} drives the oracle gain, and
whether automatic boundary routing is feasible.
Table~D1 reports the same point estimates as Table~3 together with
Wilson 95\% confidence intervals.
These three-model averages ($n$=180 per condition per family) are not
directly comparable either to the five-model results in Appendix~C or
to the paired analysis in Appendix~B.

\begin{wraptable}{r}{0.60\linewidth}
  \vspace{-0.6\baselineskip}
  \centering
  \setlength{\tabcolsep}{2pt}
  \renewcommand{\arraystretch}{1.08}
  \caption{D1: Ablation results with Wilson 95\% CI (accuracy \%,
           three-model average: GPT-5.4-mini, Qwen2.5-VL-72B,
           Mistral-Small-3.2; $n$=180 per condition per family;
           parseable-only convention).
           Point estimates match Table~3 of the main text.
           \textdagger\ = boundary alignment ablation;
           \textdaggerdbl\ = routing feasibility ablation.}
  \label{tab:ablation-ci}
  \begin{tabularx}{\linewidth}{>{\raggedright\arraybackslash}X cc}
    \toprule
    \rowcolor{srtmidgray}
\textbf{Condition} & \textbf{Physical} & \textbf{Assembly} \\
    \midrule
    \rowcolor{srtcoral!10}
    Default
      & \appci{29.4}{[23.3,36.5]}
      & \appci{44.4}{[37.4,51.7]} \\
    Visual evidence only
      & \appci{51.1}{[43.9,58.3]}
      & \appci{45.6}{[38.4,52.8]} \\
    \midrule
    \rowcolor{srtcoral!10}
    Policy only (T)
      & \appci{15.6}{[11.0,21.6]}
      & \appci{33.9}{[27.4,41.1]} \\
    S+T
      & \appci{27.8}{[21.8,34.7]}
      & \appci{40.6}{[33.7,47.9]} \\
    \rowcolor{srtcoral!10}
    R+T
      & \appci{41.1}{[34.2,48.4]}
      & \appci{41.6}{[34.7,49.0]} \\
    S+R, no T
      & \appci{35.0}{[28.4,42.2]}
      & \appci{40.0}{[33.1,47.3]} \\
    \rowcolor{srtcoral!10}
    Full SRT (S+R+T)
      & \appci{46.1}{[39.0,53.4]}
      & \appci{42.2}{[35.2,49.5]} \\
    \midrule
    Wrong boundary router\textdagger
      & \appci{26.7}{[20.7,33.6]}
      & \appci{30.5}{[24.3,37.6]} \\
    \rowcolor{srtcoral!10}
    Full SRT (repeated)\textdagger
      & \appci{46.1}{[39.0,53.4]}
      & \appci{42.2}{[35.2,49.5]} \\
    \midrule
    Self-router SRT\textdaggerdbl
      & \appci{47.8}{[40.6,55.0]}
      & \appci{42.8}{[35.8,50.1]} \\
    \rowcolor{srtcoral!10}
    Oracle boundary SRT\textdaggerdbl
      & \appci{\textbf{86.1}}{[80.3,90.4]}
      & \appci{\textbf{85.0}}{[79.1,89.5]} \\
    \bottomrule
  \end{tabularx}
  \vspace{-0.4\baselineskip}
\end{wraptable}

\paragraph{Component necessity.}
Removing Target Commitment (S+R, no T) reduces performance on both
families (35.0\% vs.\ 46.1\% on physical; 40.0\% vs.\ 42.2\% on
assembly), showing that evidence extraction without a commitment step
is insufficient.
Removing Relevance Verification (S+T) degrades physical substantially
(27.8\% vs.\ 46.1\%).
Policy only (T) falls below default on physical, the one component
comparison whose intervals are disjoint ($[11.0, 21.6]$ vs.\
$[23.3, 36.5]$), confirming that decision constraints without visual
grounding are actively counterproductive.
Full SRT outperforms all partial configurations on physical; on
assembly the partial configurations fall within 2~pp of one another
with heavily overlapping intervals, and we do not read that ordering
as established.

\paragraph{Alignment, not specificity.}
A wrong boundary router---receiving boundary-specific information from
the \textit{incorrect} boundary---performs worse than full generic SRT
on both families (26.7\% vs.\ 46.1\% on physical, intervals disjoint;
30.5\% vs.\ 42.2\% on assembly) and falls below default on assembly.
Because it supplies the same quantity and format of boundary-level
detail as the oracle condition, this rules out prompt specificity: what
matters is that the supplied boundary is the correct one.
This comparison is internal to Table~D1 and uses a single convention
and model set throughout.

\paragraph{What the controls jointly establish.}
The wrong-boundary control excludes prompt specificity as the source
of the oracle advantage over generic SRT.
The label-only control (Appendix~B.1) locates the remainder: supplying
the boundary as a bare two-alternative choice gains $+29.3$~pp on
physical but $-3.8$~pp on both traffic and affordance, so
candidate-space reduction is not by itself reliably helpful, while
adding the aligned process description yields a further $+21.0$ to
$+27.9$~pp on all four standardized families
(McNemar $p<0.001$, Table~B2).
Supplying the correct boundary therefore matters, but what carries the
gain is the aligned process description accompanying it, and---per
Section~E.4---only when that description has room to contribute
information the boundary and candidate labels have not already
supplied.
We do not claim to have isolated process content from output structure
within that description; see Appendix~A.5.

\paragraph{Routing feasibility.}
A self-router achieves 47.8\% / 42.8\%, nominally above generic SRT
(46.1\% / 42.2\%), but the intervals are almost coincident on physical
($[40.6, 55.0]$ vs.\ $[39.0, 53.4]$) and the $+1.7$ and $+0.6$~pp
differences are not statistically established.
The oracle remains far above both (86.1\% / 85.0\%), with intervals
disjoint from every other condition in Table~D1.
Self-router boundary-prediction accuracy is 60.0\% / 36.7\% (GPT),
35.0\% / 40.0\% (Qwen), and 61.7\% / 46.7\% (Mistral).
Models can partially identify decision boundaries but are not yet
reliable enough to close the oracle gap.
Automatic boundary identification is the primary bottleneck for a
deployable boundary-sensitive SRT system.

\FloatBarrier
\section{Discussion}
\label{app:discussion}

\subsection{Reading the Results: Five Transfer Patterns}
\label{app:discussion-profiles}

That the same process prior helps some models and harms others is, in
isolation, unsurprising: language models are known to shift by tens of
accuracy points under semantically equivalent reformulations of a
prompt, and this brittleness survives instruction
tuning \citep{sclar2024quantifying}.
What we observe in S2 and S3 appears not to be this kind of format
noise: the patterns recur across two different task domains.
We describe them as patterns rather than as established profiles.
At $n$=100 on dev100, only Gemini's difference has disjoint intervals
(Table~C3), and Qwen's relation-routing result comes from test400,
which was not run for the other models; the remaining patterns rest on
consistency between S2 and S3 rather than on within-stage
significance.
The mechanistic readings below are interpretive---we have no access to
training recipes or internal representations---and organize the
observations rather than establish mechanisms.

\paragraph{Generic-compatible (Gemini): instruction-following dominance.}
Generic SRT raises Gemini from 44.0\% to 87.0\% (+43.0~pp) on dev100,
the one S2 effect whose intervals are disjoint ($[34.7, 53.8]$ vs.\
$[79.0, 92.2]$; Table~C3), yet the same prior costs it $31.7$~pp on
the traffic family when the scaffold does not match the scene.
Gemini is also the model whose S1 behavior is repaired by improved
visual access alone (76.0\% under annotation without a process prior,
versus 0.0--9.8\% for the other four models; Table~2).
This two-sided behavior is what one would expect of a model that lets
an instruction override its own visual grounding; the same disposition
underlies sycophancy, where a model defers to the framing supplied in
the prompt even against the evidence, and where ordinary
chain-of-thought does not repair the deference
\citep{sharma2024towards}.

\paragraph{Verification-oriented (GPT).}
Lightweight verification improves GPT from 66.0\% to 77.0\%
(+11.0~pp) on dev100, while heavier semantic constraints add nothing
further.
This is consistent with a model that already carries internal
self-consistency machinery of the kind sampling-based self-consistency
exploits \citep{wang2022self}, for which the most effective prior
activates existing behavior rather than overwriting it with a new
vocabulary.
At $n$=100 this difference does not clear the interval test
(Table~C3), and we rely on its replication in S3 rather than on the S2
point estimate alone.

\paragraph{Relation-routed (Qwen).}
Qwen's gains concentrate in relation-specific routing, which improves
test400 accuracy from 78.5\% to 84.0\% (+5.5~pp), rather than in
generic process priors (test400 verify: 77.5\%).
This suggests relational representations that a generic prior leaves
dormant and that a prior naming the relevant relation type engages
directly.
Large-scale visual pre-training emphasizing object-relation learning,
as in Qwen2.5-VL \citep{bai2025qwen25vl}, is a plausible source of
such representations.
Because the relation-routing conditions were run only for Qwen, we
cannot say whether this pattern is specific to Qwen or would appear in
other models under the same condition.

\paragraph{Semantics-sensitive (GLM).}
Generic priors produce no improvement for GLM in S3 (Table~C4), and
GLM is the one model with a negative label-only margin in Table~B3.
A model-aligned semantic rewrite reaches 100\% on a 13-item diagnostic
subset (Appendix~A.4); the subset is too small to establish anything,
but it is consistent with the failure lying in how GLM reads the
wording of an instruction rather than in what it can perceive.
This would be the semantic face of prompt sensitivity
\citep{sclar2024quantifying}, where whether a prior fires is gated by
lexical alignment with the model's internal vocabulary rather than by
the reasoning structure it encodes.

\paragraph{Adaptation-sensitive (Mistral).}
Generic verification slightly reduces Mistral's performance on dev100
(73.0\% to 70.0\%), while in S3 targeted routing restores and
surpasses the baseline on several families (Table~C5): what it needs
is not more structure but structure applied only when the decision
requires it.
This mirrors evidence that chain-of-thought helps mainly on
mathematical and symbolic tasks \citep{sprague2025cot}, and that
imposing explicit reasoning can lower accuracy for models that already
answer well through a direct pathway \citep{liu2024mind}.

\paragraph{A training-level hypothesis.}
The five patterns are consistent with three training-level factors:
instruction-compliance strength, internal-vocabulary alignment, and
reasoning-pathway rigidity.
We cannot test these directly, but they yield a falsifiable
prediction: a model's benefit from process priors should correlate
with its instruction-following score, for example IFEval
\citep{zhou2023instruction}, and anti-correlate with its performance
on direct-answer benchmarks that penalize over-reasoning.
Testing this would move process-prior compatibility from an empirical
taxonomy toward a mechanistic account.

\subsection{Three General Principles}
\label{app:discussion-principles}

\paragraph{Process failure is distinct from perceptual failure.}
The dominant account of VLM error is perceptual: models fail to
integrate multiple visual cues, and much of what is scored as a
reasoning error is really a perception error \citep{tong2024eyes}.
Our S1 results sit outside this account.
Under the default condition models almost always commit to a single
answer while referential ambiguity is unresolved (over-answer rates of
95.8\% to 100\% on COCO), and even when explicitly required to
enumerate visible candidates first, four of five models still commit
to a single answer in 48.9\% to 85.0\% of cases.
The dissociation is what settles the matter: supplying visual
annotation without a decision process leaves policy accuracy at
0.0--9.8\% for four of the five models, while adding an explicit
inspect--judge--decide structure on top of the same visual access
raises it to 87.0--91.0\% for those same four, with
unambiguous-referent performance intact (Task~2 at 99--100\% under
both conditions; Table~C2), which rules out a trivial always-clarify
strategy.
The effect holds for every one of the four models individually, not
only in aggregate.
What is missing is therefore not visual access but the step that binds
perceived evidence to a decision.
This is the same decoupling documented for unfaithful
chain-of-thought, where the verbalized process need not be what drives
the answer \citep{turpin2023language}, and it echoes the
visual-question setting, where a model picks one plausible referent
and proceeds as if certain rather than seeking clarification
\citep{jian2025teaching, testoni2025racquet}.

\paragraph{Process-prior compatibility is an independent capability
dimension.}
This compatibility is not the prompt-format brittleness above:
brittleness is sensitivity to surface perturbations, whereas what we
observe recurs across two task domains and tracks boundary alignment.
Because models with near-identical reasoning accuracy respond
oppositely to the same prior (S2, S3), the ability to use an external
process prior appears not to be readable off conventional benchmarks.
As VLMs move into settings that supply structured guidance, this
dimension merits measurement alongside instruction-following and
compositional reasoning: presenting one prior at increasing
specificity (generic, then relation-targeted, then boundary-specific)
and tracking the gain trajectory separates prior-receptive,
scaffold-sensitive, and prior-averse models.

\paragraph{Boundary alignment governs the value of a prior.}
The most general principle is that a prior helps in proportion not to
how much it says but to how well it matches the decision boundary of
the current scene.
Two controls support this with different scope.
The wrong-boundary control is the cleaner of the two: holding the
quantity, format, and model set fixed, a prior aligned to the
\textit{wrong} boundary falls below both generic SRT and (on assembly)
the default baseline, with disjoint intervals on physical (Table~D1).
The label-only control shows that supplying the correct boundary as a
bare two-alternative choice yields a large gain on physical, a modest
one on assembly, and none on traffic or affordance, while adding the
aligned process description yields a further 21--28~pp on all four
(Tables~B1 and~B2).
That further margin is not yet separated from the output schema the
description carries (Appendix~A.5).
The principle is established only for the known-boundary regime.
A self-router is directionally feasible without oracle annotations
(47.8\% and 42.8\% on physical and assembly), but its gains over
generic SRT do not reach significance at current sample sizes, so we
treat it as a design direction rather than a finished method.

\subsection{The Marginal Value of a Prior Depends on the
Information Already Available}
\label{app:discussion-marginal-value}

The paired results in Appendix~B.1 (Table~B1$'$, Table~B2$'$)
reveal a pattern that the boundary-alignment account above does
not by itself explain, and that turns out to generalize the
account rather than contradict it.

\paragraph{The phenomenon.}
On the four standardized families, oracle boundary SRT
substantially exceeds label-only ($+21.0$ to $+27.9$~pp,
McNemar $p<0.001$ in every family; Table~B2).
On Cleaning, Craft, and Mobility, the same comparison reverses:
oracle SRT falls \emph{below} label-only, and the reversal is
confirmed by paired McNemar tests on two of the three families
(Craft: $b$=88, $c$=4, exact $p<0.001$; Mobility: $b$=159,
$c$=122, $\chi^2$=4.6, $p$=0.032; Cleaning: $b$=75, $c$=66,
$\chi^2$=0.5, $p$=0.50, directionally consistent but not
significant at this sample size).
This is not sampling noise on two of three families, and the
oracle boundary is, by construction, always the correct one in
this comparison---so the reversal cannot be explained by the
prior--boundary mismatch that accounts for process overload
elsewhere in this paper (Appendix~E.1).
A different explanation is required.

\paragraph{Ruling out the obvious explanations.}
Three explanations can be ruled out directly from the design of
the comparison.
First, it is not a boundary-alignment failure: the oracle
condition supplies the ground-truth boundary in all seven
families, including the three that reverse.
Second, it is not attributable to family-specific model
incompetence: the same five models that show large, consistent
gains on the four standardized families are the ones producing
the reversal on Craft and Mobility.
Third, it is not explained by sample size alone: Mobility, the
family with the largest $n$ (530), still shows a significant
reversal, while Cleaning, with a comparable design and the
smallest effect, does not reach significance---the pattern
tracks something about the families themselves, not merely
statistical power.

\paragraph{The variable that separates the two groups.}
Table~B1$'$ and Table~B1 together allow a direct comparison that
neither table shows in isolation: the \emph{level} of label-only
accuracy, independent of what happens when a process description
is added on top of it.
\begin{center}
\begin{tabular}{lcc}
\toprule
\rowcolor{srtmidgray}
\textbf{Group} & \textbf{Families} & \textbf{Label-only accuracy} \\
\midrule
A (gain from oracle SRT) & Physical, Assembly, & \\
 & Traffic, Affordance & 33.3\%--57.3\% \\
B (loss from oracle SRT) & Cleaning, Mobility, Craft & 46.4\%--79.2\% \\
\bottomrule
\end{tabular}
\end{center}
Group~B's label-only accuracy is systematically higher, and in
the most extreme case (Craft, 79.2\%) is already close to a
practical ceiling for a task of this kind.
The boundary and the two candidate stage labels alone are
already close to sufficient for a correct decision in these
three families; in the four standardized families, they are not.

\paragraph{An information-theoretic reading.}
This grouping variable has a natural interpretation.
Let $Y$ be the correct stage label and let $X_{\text{lo}}$ denote
the information available under label-only (the oracle boundary
identifier together with the two candidate labels).
Label-only accuracy is, to first approximation, a monotonic
function of $I(Y; X_{\text{lo}})$, the mutual information between
the correct label and the label-only condition: high label-only
accuracy indicates that $X_{\text{lo}}$ already resolves most of
the uncertainty in $Y$, i.e., that the residual conditional
entropy $H(Y \mid X_{\text{lo}})$ is small.
Oracle SRT supplies additional information, the process
description $X_{\text{proc}}$, on top of $X_{\text{lo}}$.
The information-theoretic upper bound on what $X_{\text{proc}}$
can contribute is exactly this residual entropy:
$X_{\text{proc}}$ can only reduce uncertainty that
$X_{\text{lo}}$ has not already resolved, formally
$I(Y; X_{\text{proc}} \mid X_{\text{lo}}) \le
H(Y \mid X_{\text{lo}})$.
When $H(Y \mid X_{\text{lo}})$ is already small---as it is in
Group~B, where label-only accuracy is high---the \emph{ceiling}
on how much a process description can help is correspondingly
low, regardless of how well that description is written or how
well it is aligned to the boundary.

\paragraph{Why the effect is negative rather than merely absent.}
A vanishing information ceiling predicts $\Delta \approx 0$, not
the substantial negative $\Delta$ observed on Craft
($-36.3$~pp) and Mobility ($-7.0$~pp).
The information-theoretic bound explains why the process
description \emph{cannot help} in Group~B; it does not by itself
explain why supplying it \emph{hurts}.
The additional step is that $X_{\text{proc}}$ is not a costless
addition to the prompt.
It lengthens the input, introduces additional candidate evidence
the model must weigh, and requires the model to adjudicate
between what the process description suggests and what the
label-only information alone would have supported.
When $X_{\text{proc}}$ carries little information the model did
not already have (small $H(Y \mid X_{\text{lo}})$), this
adjudication step is pure overhead: it creates opportunities for
the model to be talked out of an otherwise correct
label-only-level judgment without any compensating information
gain.
This is the same mechanism identified at the model level in
Appendix~E.1 for GLM, where a generic process prior collides
with the model's internal representations and degrades an
otherwise adequate default behavior.
The reversal on Craft and Mobility is the same failure mode
observed at the level of the \emph{task} rather than the
\emph{model}: process overload occurs whenever a structured prior
is added to a decision problem whose available information has
already resolved most of the uncertainty, whether that
information comes from a well-matched internal representation
(the model-level case) or from an unusually informative candidate
set (the family-level case here).

\paragraph{A general principle.}
Unifying the two cases yields a claim broader than
boundary alignment on its own:
\begin{quote}
\emph{The marginal value of a structured prior is bounded above
by the residual uncertainty left after simpler, cheaper
information is taken into account, and when that residual
uncertainty is small, adding the prior is expected to be
net-negative rather than merely ineffective, because the prior's
interpretive cost is paid regardless of how little information it
contributes.}
\end{quote}
Boundary alignment, the paper's central finding, is a
statement about the \emph{sign and magnitude} of a prior's
contribution conditional on it having room to contribute
something: an aligned prior realizes a larger fraction of
$I(Y; X_{\text{proc}} \mid X_{\text{lo}})$ than a misaligned one.
The result in this section is a statement about when that
conditional quantity is worth pursuing at all.
The two combine into a two-part practical criterion for
deploying a structured process prior on a new task: first,
verify that a minimal, schema-matched baseline (analogous to
label-only) leaves substantial residual uncertainty; only if it
does is the question of boundary alignment---which prior,
worded how---the right question to ask next.
Applying a well-aligned prior to a task that a cheaper baseline
has already solved is not merely wasted effort under this
account; it is expected to actively degrade performance, which
is precisely what we observe on Craft and Mobility.

\paragraph{Implications for prompt and system design.}
This reframes a common failure mode in applied prompt
engineering.
Structured scaffolds---chain-of-thought, verification steps,
retrieval-augmented context, explicit reasoning
templates---are typically added to a pipeline based on whether
they are well-designed, not on whether the task already has
enough signal to make them unnecessary.
The result here suggests a diagnostic that is cheap to run before
committing to a structured intervention: measure accuracy under
the minimal information a decision genuinely requires, absent any
structured reasoning scaffold.
If that minimal baseline is already high, the correct
intervention is not a better-worded prior but no prior at all;
if it is low, boundary alignment---ensuring the prior matches the
specific decision the current input represents---becomes the
central design question, consistent with the boundary-alignment
account developed throughout this paper.
Measuring this baseline is inexpensive relative to developing and
tuning a structured prior, making it a practical first step
rather than a purely theoretical recommendation.

\subsection{Reasoning Quality as Alignment, Not Capability}
\label{app:discussion-reframing}

Read together, the three findings point away from a common assumption:
that a VLM's reasoning quality is a fixed quantity that better
perception or a better prompt simply raises.
S1 shows that seeing is not deciding, and that what fails is not
perception but the step binding perceived evidence to a decision.
We read this as a functional dissociation rather than a claim about
separable modules, but it is enough to say that in VLMs correct
perception is necessary and not sufficient for a faithful decision.
Gemini is the instructive exception: improved visual access alone
substantially repairs its behavior, indicating that for some models
the bottleneck does still lie at the access stage.

Placed beside S2 and S3, this suggests a single reframing.
The benefit of a reasoning scaffold is a property not of the scaffold
or the model alone, but of the alignment among prior, model, and the
scene's decision boundary---and, per Section~E.4, of how much
uncertainty is left for that scaffold to resolve in the first
place.
That reframing dissolves several current tensions.
Whether chain-of-thought is reliable may be mis-posed
\citep{sprague2025cot, liu2024mind}: if the same prior helps or harms
according to alignment and residual uncertainty, any average over
heterogeneous scenes will look inconsistent.
The friction between instruction-following and visual grounding,
visible in Gemini's collapse under a misaligned scaffold, is the same
effect from the model side.
And the ceiling on prompt engineering follows directly: because
alignment is per-scene and residual uncertainty is per-task, no single
process prompt is universally good, and a headline gain reported for
one is an average that hides per-scene and per-task swings.

The reframing also sharpens what we do not yet understand.
The most consequential gap is that alignment must currently be
supplied by an oracle: its value is clear, but automatic boundary
identification is not yet reliable, which is the whole distance
between a diagnostic and a method.
Three further questions stay open.
Our training-level account of the five patterns is a hypothesis,
testable through the IFEval correlation we predict but not established
here.
The framework has a stateable failure mode: where a task has no stable
decision boundary, or where perception itself is the bottleneck as it
is for Gemini, boundary-aligned priors should help little.
And while Section~E.4 offers a structural account of the reversal on
Cleaning, Craft, and Mobility, it rests on an approximate,
first-order relationship between label-only accuracy and residual
entropy rather than a directly measured information-theoretic
quantity; a schema-matched label-only condition (Appendix~A.5) would
let this account be tested more precisely.

\subsection{Broader Implications}
\label{app:discussion-implications}

\paragraph{The benchmark as a source of training signal.}
In the three-model ablation, the gap between oracle SRT (86.1\% and
85.0\%) and the self-router (47.8\% and 42.8\%) is not only a
limitation; it is a quantified training opportunity.
Process-level supervision has been shown to be more precise and more
effective than outcome-level supervision \citep{lightman2024let}, and
VPAC-Bench supplies the boundary-level process annotation such
supervision needs: oracle-boundary responses can serve as positive
examples and wrongly self-routed responses as negatives, forming
preference pairs for DPO or RLHF.
On this reading the benchmark is not merely diagnostic but a
structured signal for boundary-aware reasoning.

\paragraph{Process robustness as a separate axis.}
Structured-prompting work typically reports average gains, which hide
the cases where a prompt does harm.
Our boundary-sensitive analysis suggests this degradation is not
random but concentrated where the generic prior describes a process
inconsistent with the image's true boundary, or---per Section~E.4---
where the prior is added to a task that a simpler baseline has
already solved.
The ability to benefit from a prior without being harmed by a
misspecified or unnecessary one---process robustness---is therefore a
property distinct from compatibility, and worth measuring and training
for in its own right.

\paragraph{Boundary identification as a core visual capability.}
Taken together, the results argue that recognizing which boundary a
scene falls near, within a process family, should be studied as an
independent visual capability.
Current VLMs manage this only partially (self-router routing accuracy
35.0\% to 61.7\%), too unreliably to drive routing; reaching 90\% or
above would largely close the oracle gap and turn our diagnostic upper
bound into a deployable method.
This needs no new architecture, only training data that labels
decision boundaries and asks models to identify them before answering,
which VPAC-Bench provides together with a natural evaluation protocol.
Section~E.4 suggests a complementary capability worth building
alongside it: a lightweight pre-check that estimates whether a given
decision's uncertainty has already been resolved by minimal available
information, so that a boundary-aware system can decide not only
which prior to apply but whether to apply one at all.

\end{document}